\documentclass[10pt,twocolumn,letterpaper]{article}

\usepackage[pagenumbers]{cvpr} 

\usepackage{xcolor}
\definecolor{darkgreen}{rgb}{0.0, 0.5, 0.0}
\definecolor{darkred}{rgb}{0.6, 0.0, 0.0}

\newcommand{\appx}{Appendix}

\newcommand{\coco}{\texttt{COCO}}
\newcommand{\flickr}{\texttt{Flickr30k}}
\newcommand{\itw}{\texttt{In-the-wild}}
\newcommand{\syn}{\texttt{Synthbuster}}
\newcommand{\glvtwo}{\texttt{GLDv2}}
\newcommand{\glide}{\texttt{GLIDE}}
\newcommand{\dalle}{\texttt{dalle3-images}}
\newcommand{\realsdxl}{\texttt{realisticSDXL}}
\newcommand{\ddb}{\texttt{diffusiondb}}
\newcommand{\mji}{\texttt{mj-images}}
\newcommand{\mjtti}{\texttt{mj-tti}}
\newcommand{\sdone}{\texttt{SD1.X}}
\newcommand{\sdthree}{\texttt{SD3}}
\newcommand{\sdtwo}{\texttt{SD2}}
\newcommand{\sdonethree}{\texttt{SD1.3}}
\newcommand{\sdonefour}{\texttt{SD1.4}}

\newcommand{\dalletwo}{\texttt{DALLE2}}
\newcommand{\dallethree}{\texttt{DALLE3}}
\newcommand{\mjearly}{\texttt{MJ 1/2}}
\newcommand{\mjlate}{\texttt{MJ 5/6}}
\newcommand{\sdxl}{\texttt{SDXL}}
\newcommand{\flux}{\texttt{FLUX.1}}
\newcommand{\firefly}{\texttt{Firefly}}
\newcommand{\susy}{\texttt{SuSy}}
\newcommand{\thesusy}{\texttt{SuSy}}
\newcommand{\justsusy}{\texttt{SuSy}}

\definecolor{cvprblue}{rgb}{0.21,0.49,0.74}
\definecolor{lightgray}{gray}{0.9} 
\usepackage[pagebackref,breaklinks,colorlinks,allcolors=cvprblue]{hyperref}
\usepackage{placeins}
\usepackage{arydshln}
\usepackage{array}
\usepackage{booktabs}
\usepackage{colortbl}
\usepackage{xcolor} 

\definecolor{lightorange}{RGB}{255, 230, 180} 
\definecolor{lightblue}{RGB}{224, 234, 247}   

\usepackage[status=draft,nomargin,inline,lang=spanish]{fixme}
\fxusetheme{colorsig}
\FXRegisterAuthor{dg}{adg}{}

\def\paperID{16701}
\def\confName{CVPR}
\def\confYear{2025}

\title{Present and Future Generalization of Synthetic Image Detectors}

\author{Pablo Bernabeu-Pérez\\
Barcelona Supercomputing Center\\
{\tt\small pablo.bernabeu@bsc.es}
\and
Enrique Lopez-Cuena\\
Barcelona Supercomputing Center\\
{\tt\small enrique.lopez@bsc.es}
\and
Dario Garcia-Gasulla\\
Barcelona Supercomputing Center\\
{\tt\small dario.garcia@bsc.es}
}

\begin{document}
\maketitle

\begin{abstract}

The continued release of increasingly realistic image generation models creates a demand for synthetic image detectors. To build effective detectors we must first understand how factors like data source diversity, training methodologies and image alterations affect their generalization capabilities. This work conducts a systematic analysis and uses its insights to develop practical guidelines for training robust synthetic image detectors. Model generalization capabilities are evaluated across different setups (\eg scale, sources, transformations) including real-world deployment conditions. Through an extensive benchmarking of state-of-the-art detectors across diverse and recent datasets, we show that while current approaches excel in specific scenarios, no single detector achieves universal effectiveness. Critical flaws are identified in detectors, and workarounds are proposed to enable the deployment of real-world detector applications enhancing accuracy, reliability and robustness beyond the limitations of current systems. 

\end{abstract}

\section{Introduction}

Synthetic image generation is presenting challenges regarding visual information integrity, mitigation of misinformation, and, trust and rights in digital environments. Due to these concerns, correctly attributing synthetic content has become a social demand and a top scientific priority. Recent legislation aligns with this context, mandating the identification and notification of synthetic digital content~\cite{ai-act}. 




To address these needs, synthetic image \textit{detection} (SID) has become locked in a race with synthetic image \textit{generation} (SIG)~\cite{laurier2024cat}. SID aspires to win by developing universal detectors~\cite{ojha2023towards,chendrct}, but their generalization capacity remains uncertain. Meanwhile, new SIG models join the race every month, moving forward in realism and posing new challenges to SID models. This work studies the relation between SIG and SID, by first analyzing the impact of training conditions on SID generalization (\S\ref{sec:train_experiments}). The learnt lessons are applied to train a baseline for evaluating the generalization capacity on deployment conditions. This includes variations in data and model source (who used the SIG and which SIG was used) and scaling factors (\S\ref{sec:eval}). The last set of experiments (\S\ref{sec:generalization_sota}) include an updated benchmark on recent detectors, using synthetic data produced by the latest generators, under an optimized image scaling policy. Finally, the ethical considerations related to SID research and development, including when and how should detectors be publicly released, are discussed (\S\ref{sec:concl}).

Findings indicate current methods are insufficient for reliable SID, as no tested model generalizes universally. Factors like rescaling play a major role in detector performance, exposing a vector of attack for malicious actors. While some models suffer major degradations, others benefit from a resized input, emphasizing the importance of choosing the right preprocessing techniques. Lastly, detectors perform much worse on private models, like \texttt{DALLE} and \texttt{Midjourney}, compared to open models, highlighting the crucial role of open science for synthetic attribution. This work illustrates how, as of today, generalization should never be assumed in the field of SID.






\section{Related Work}


Previous work on SID has largely focused on GAN-generated content~\cite{zhang2019detecting,marra2019gans,wang2020cnn,chai2020makes,frank2020leveraging,giudice2021fighting}, primarily due to their historical prevalence and relative speed. However, recent studies reveal that GAN-focused detectors often fail to identify content from modern diffusion models ~\cite{wang2023dire,lorenz2023detecting}.  While several recent works have addressed the detection of diffusion-based content~\cite{lin2024robust,aghasanli2023interpretable,wissmann2024whodunit,corvi2023detection,xu2023exposing,liu2024forgery,zhu2024genimage,zhu2023gendet,grommelt2024fake,ojha2023towards}, which now produces the most perceptually convincing synthetic images, their generalization ability under different conditions remains mostly untested.

Across all families of detection methods, frequency domain-based approaches are commonly used to detect synthetic content, revealing generation artifacts~\cite{chandrasegaran2021closer,corvi2023intriguing}. Some methods leverage Fast Fourier Transform analysis to capture characteristic patterns~\cite{tan2024frequency,bammey2023synthbuster}, while another recent work~\cite{deng2023diffusion} explores wavelet-based features specifically tailored for diffusion outputs. Deep learning architectures such as CNNs ~\cite{coccomini2024detecting,sinitsa2024deep,tan2023learning} and Visual Transformers (ViTs) ~\cite{aghasanli2023interpretable,liu2024forgery} have been used to learn hierarchical synthetic patterns, with CLIP-based methods further enhancing detection through semantic~\cite{ojha2023towards} and intermediate feature analysis~\cite{koutlis2024leveraging}. Models combining textual and visual features have also been adopted for SID; while~\cite{chang2023antifakeprompt} applies prompt tuning to detect deepfakes, by approaching detection as a visual question-answering problem, \cite{wu2023generalizable} performs contrastive learning via text guidance. Hybrid models combine multiple detection signals to improve generalization, such as dual-stream networks analyzing texture and frequency artifacts~\cite{xi2023ai} and CLIP features fused with low-level image statistics~\cite{song2024trinity}. Finally, local feature analysis is used to examine texture contrast patterns~\cite{zhong2024patchcraft} and intrinsic dimensionality properties~\cite{lorenz2023detecting} to provide complementary signals.

AI-generated image detector models are usually trained using data from a single source and evaluated on datasets from multiple sources to assess their generalization capacity, ~\cite{corvi2023detection,ricker2022towards,zhu2024genimage,zhu2023gendet,grommelt2024fake,ojha2023towards,cazenavette2024fakeinversion,chendrct}. Among the various sources of bias that have been examined, image format and resolution stand out as key factors. In~\cite{corvi2023detection}, authors highlight the impact of the resizing operation, a common practice in deep learning to adjust images to the network's input resolution. The study presented in~\cite{grommelt2024fake} highlights biases associated with \textit{JPEG} compression and image size. Authors demonstrate a size bias affecting detector performance, with detectors generally performing better on natural images that differ significantly in size from the generated images used in training. This observation aligns with findings in~\cite{cozzolino2024raising}, where it is demonstrated that dataset choice significantly impacts detection performance. Meanwhile, other relevant factors for generalization remain to be studied, including model family, model release date, and dataset source. 

\section{Methods}

To examine detector biases arising from training methodology, we employ a fixed architecture (see~\S\ref{subsec:arch}), train it using six image datasets (see~\S\ref{subsec:data}) and evaluate with fifteen additional datasets (see~\S\ref{subsec:data_eval}). To enable full reproducibility of the work, our codebase~\footnote{https://github.com/HPAI-BSC/SuSy}, training datasets~\footnote{https://huggingface.co/datasets/HPAI-BSC/SuSy-Dataset} and model weights~\footnote{https://huggingface.co/HPAI-BSC/SuSy} for our best detector are publicly released.

\subsection{Architecture}\label{subsec:arch}

The two popular architectural choices for building a SID are training a direct classifier, or using the features extracted from a pre-trained model. Both CNNs~\cite{ricker2022towards,zhu2024genimage,grommelt2024fake} and ViTs~\cite{xu2023exposing, zhu2023gendet,ojha2023towards,cozzolino2024raising} have been considered for those purposes, both of them performing competitively. For our experimentation, we choose a ResNet~\cite{he2016deep} trained as a direct classifier, as this has shown competitive and robust results~\cite{zhu2024genimage,grommelt2024fake,cazenavette2024fakeinversion} while being a lightweight architecture, allowing evaluation at scale. The staircase design proposed in~\cite{lopez2023sr} is used (12.7M parameters), which feeds features extracted at different blocks into a multi-layer perceptron (see \appx~\ref{appendix-arch}).

Detectors are commonly trained either on image patches or on downsampled images, as processing entire high-resolution images is computationally intensive and the most discriminating features are typically low-level. For each image, we select five 224$\times$224 patches exhibiting the highest contrast in their grey-level co-occurrence matrix~\cite{haralick1973textural}. These patches are individually processed through the network, producing per-patch predictions that must be aggregated to obtain image-level decisions. Various combination strategies and their impact on detection performance are studied in \S\ref{subsec:image_eval}.

Regardless of evaluating at patch or image level, detection performance metrics are chosen based on the number of datasets being analyzed. We employ recall when evaluating performance on a single dataset (either authentic or synthetic), focusing exclusively on the model's effectiveness in identifying the class at hand. This avoids misleading interpretations that could arise from metrics considering both positive and negative classes. For multi-dataset classification scenarios, we utilize macro accuracy, which provides an unweighted mean of per-class accuracy, ensuring fair evaluation across all classes regardless of sample size.

\subsection{Train Datasets}\label{subsec:data}

The training experiments detailed in \S\ref{sec:train_experiments} utilize two types of datasets: \textit{authentic} real-world images sourced from \coco~\cite{lin2015microsoft} and \textit{synthetic} AI-generated images from \dallethree{}~\cite{dalle-3-images}, \sdone{}~\cite{wangDiffusionDBLargescalePrompt2022}, \sdxl{}~\cite{realisticSDXL}, \mjearly{}~\cite{iulia_turc_gaurav_nemade_2022} and \mjlate{}~\cite{midjourney-images}. These represent different versions of the three most popular image generators: \texttt{DALLE}, \texttt{StableDiffusion} and \texttt{Midjourney}. To ensure balanced class representation, \coco{} and \sdone{} are undersampled to a maximum of 5,435 images. The pre-existing train, validation and test splits are respected, defaulting to a standard 60\%-20\%-20\% random split distribution when such partitions are not available. For the \sdxl{} dataset, the \textit{realistic-2.2} split was used for training and validation purposes, while the \textit{realistic-1} split was used for testing. Further details regarding release dates, image formats, resolutions and dataset split sizes are available in \appx~\ref{appendix-train-data}.

\subsection{Benchmarking Datasets}\label{subsec:data_eval}

To evaluate SID models we use fifteen datasets: eleven produced and gathered by others, two produced by others but gathered by us, and two produced by us. Image resolution distributions and visual samples are provided in Appendices~\ref{appendix-resolution} and~\ref{appendix-sample-images}, respectively.

The datasets produced by others include two subsets of 5,000 randomly selected authentic images: scenes depicting people from \flickr{}~\cite{young2014image} and natural and human-made landmarks from \glvtwo{}~\cite{weyand2020google}. Additionally, nine synthetic datasets from the \syn{} superset~\cite{bammey2023synthbuster} provide 1,000 images each, generated using a common set of prompts across both models included in our training (\eg \sdxl{}, \dallethree{}) and models outside our training set (\eg \dalletwo{}, \firefly{}).

\begin{figure}[t]
  \centering
  \includegraphics[width=0.9\linewidth]{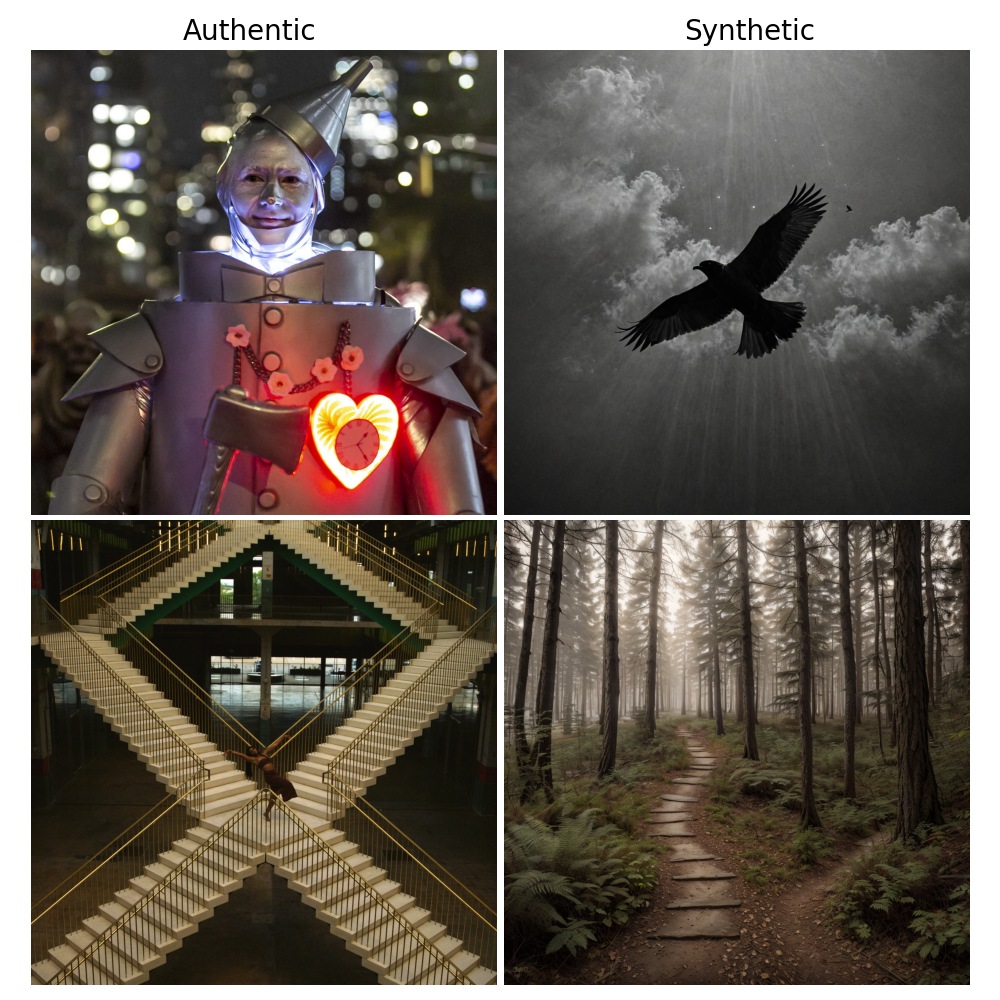}
  \caption{Examples of the \itw{}~dataset.}
  \label{fig:itw}
\end{figure}

The \itw{} dataset, as shown in Figure~\ref{fig:itw}, comprises both authentic and synthetic images gathered from online sources by the authors. The authentic split contains 121 images manually collected from \textit{Reddit} (from communities that forbid AI content) and \textit{Flickr} (from uploads prior to 2020), while the synthetic split consists of 99 photorealistic AI-generated images sourced from \textit{Civitai} and \textit{Reddit}'s synthetic content communities.

Finally, we generate two 8,192 synthetic image datasets: \sdthree{}, generated using Stable Diffusion 3-Medium~\cite{StabilityAI2024SD3}, a MMDiT text-to-image model, and~\flux{}, created with \texttt{FLUX.1-dev}~\cite{BlackForestLabs2024}, a 12B parameter model that combines MMDiT and DiT~\cite{li2022dit} architectures. Additional details are provided in \appx~\ref{appendix-generated}.

\section{Train Experiments}\label{sec:train_experiments}


This section examines how different training strategies affect model generalization. For consistent experimental comparison, all models share identical architecture (\S\ref{subsec:arch}) and hardware setup (\appx~\ref{appendix-setup}). The training is capped at 20 epochs with a 2-epoch patience early stopping based on validation accuracy. The datasets described in \S\ref{subsec:data} are augmented using horizontal flips with 50\% probability, while additional transformations are analyzed in \S~\ref{subsec:alterations}.

\subsection{Single-class Models}\label{sec:single-class}


We evaluate relationships between SIG models by training binary classifiers, using each synthetic dataset in \S\ref{subsec:data} as a positive class and \coco~as the negative class. These single-class detectors are then tested on the remaining datasets to assess cross-model generalization (see Table~\ref{tab:single_class_recall_patch}).

\begin{table}[t]
\setlength\tabcolsep{2.4pt}
\centering
\small
\begin{tabular}{c|ccccc|c}
                     &\dallethree & \sdone & \sdxl & \mjearly & \mjlate & Avg. \\ 
                     \hline
\ \dallethree & \textbf{97.70} & 27.59          & 70.19          & 50.73          & 97.02          & 68.64 \\
\ \sdone      & 49.76          & \textbf{98.30} & 68.23          & 39.65          & 40.36          & 59.26 \\
\ \sdxl       & 51.27          & 33.45          & \textbf{97.57} & 59.14          & 67.97          & 61.88 \\
\ \mjearly    & 31.39          & 17.63          & 73.14          & \textbf{99.07} & 51.51          & 54.55 \\
\ \mjlate     & 91.76          & 26.13          & 64.75          & 62.69          & \textbf{99.25} & 68.92 \\ \hline
Avg.          & 64.38          & 40.62          & 74.78          & 62.26          & 71.22          &    
\end{tabular}
\caption{On each row, patch-level recall of single-class models for synthetic datasets. In bold, performance on the training dataset.}
\label{tab:single_class_recall_patch}
\end{table}


While single-class detectors achieve excellent recall (over 97\%) on their target class, performance drops substantially when tested on other datasets. SIG model age emerges as the dominant factor affecting generalization performance. When evaluating newer detectors on older generators, we observe severe performance degradation, as shown in the last row of Table~\ref{tab:single_class_recall_patch}, where detectors trained on \sdone{} and \mjearly{} (both from 2022) show the lowest average values. This pattern likely stems from older generators producing more pronounced artifacts, which newer detectors struggle to identify without specific training. Conversely, detectors trained on recent datasets show better cross-SIG generalization, as evidenced by the higher average values for \dallethree{}, \sdxl{} and \mjlate{} in the last column. Paradoxically, this suggests that more realistic generators enhance the robustness and reduce the bias of detectors. In contrast, SIG family has a weak effect on generalization. The detector trained on \sdxl{} is below average when tested on \sdone{}. Likewise, the SID model trained on \mjearly{} is not particularly accurate on \mjlate{}. The effect of image format is also inconclusive.


\subsection{Multi-class Models}


Multi-class detectors offer richer decision boundaries compared to single-class detectors, which tend to collapse~\cite{del2022multiclass} \ie defaulting to predicting only one class. To explore the effects of this distinction on generalization, we train a binary classifier merging all synthetic data sources from~\S\ref{subsec:data} into a single synthetic class, including 14,323 synthetic images and an analogous amount drawn from~\coco{} to compose the authentic class. We also train a six-class recognition model using the original splits defined in~\S\ref{subsec:data}. To obtain binary classifications from the six-class model, we take \textit{argmax} of the output probabilities, where all samples labeled as belonging to a synthetic class are considered equal predictions of the synthetic class. An alternative threshold mechanism was explored, with minimal impact on performance, and its results are reported in Appendix~\ref{appendix-threshold}.


\begin{table}[t]
\setlength\tabcolsep{1.8pt}
\centering
\begin{tabular}{c|cccccc}
        & Auth.          & \dallethree    & \sdone         & \sdxl          & \mjearly       & \mjlate        \\ \hline
Single  & -              & 97.70          & 98.30          & 97.57          & 99.07          & 99.25          \\
Binary  & 94.85          & 99.64          & \textbf{98.98} & 99.22          & 99.60          & 99.74          \\
6 Class & \textbf{97.39} & \textbf{99.76} & 97.89          & \textbf{99.30} & \textbf{99.91} & \textbf{99.97} \\
\end{tabular}
\caption{Patch level recall for \textit{Single}: five models trained on each dataset separately (\ie Table~\ref{tab:single_class_recall_patch} diagonal). \textit{Binary}: model trained with all synthetic datasets merged. \textit{Six-class}: Multi-class model trained for the recognition task. Best in bold.}
\label{tab:multi_class_confusion}
\end{table}

Results in Table~\ref{tab:multi_class_confusion} show a good performance from both the binary and the six-way classifiers on all synthetic datasets. Better than single models, which means visual features of synthetic detectors are mutually beneficial for SID. In general, the six-way classifier outperforms all, with the only exception of one of the oldest and most distinct datasets (\ie \sdone) (lowest generalization in Table~\ref{tab:single_class_recall_patch}).

\subsection{Image Alteration Methods} \label{subsec:alterations}



Image transformations, while essential for storage optimization and transmission cost reduction, can significantly alter images and may be exploited by malicious actors to mask synthetic content. If image analysis models are not robust to these transformations, their utility in real-world scenarios becomes minimal. To evaluate this robustness, we test the six-class model from the previous section under several common transformations from the Albumentations library~\cite{info11020125}: blur (\textit{AdvancedBlur} and \textit{GaussianBlur}), brightness and gamma alterations (\textit{RandomBrightnessContrast} and \textit{RandomGamma}), and \textit{JPEG compression}, all using default parameters.

For a complete assessment, we train five multi-class models, each with a different transformation applied to its training set, and evaluate these alongside our original six-class model across all transformations and unaltered images. The results are presented in Table~\ref{tab:alteration_confusion} using multi-class macro accuracy, where any misclassification between synthetic classes is counted as an error. This metric was selected instead of the previously used binary metrics, as binary classification consistently achieved over 99\% accuracy, limiting its ability to distinguish model performance in a multi-class context.


\begin{table}[tb]
\setlength\tabcolsep{2.8pt}
\centering
\begin{tabular}{c|cccccc|c}
         & None & Bright & $\gamma$ & JPEG & ABlur & GBlur & Avg. \\
\hline
6 Class & \textbf{90.90} & 86.66          & 90.60          & 90.19          & 81.56          & 54.73          & 82.44 \\
Bright   & 91.28          & \textbf{89.68} & 91.13          & 90.10          & 84.61          & 63.55          & 85.06 \\
$\gamma$ & 91.52          & 87.51          & \textbf{91.30} & 90.02          & 85.57          & 65.22          & 85.19 \\
JPEG     & 87.82          & 83.15          & 87.79          & \textbf{86.21} & 78.42          & 55.29          & 79.78 \\
ABlur    & 90.13          & 86.23          & 90.12          & 88.15          & \textbf{88.04} & 81.54          & 87.37 \\
GBlur    & 88.94          & 84.02          & 88.65          & 87.37          & 86.78          & \textbf{81.88} & 86.27 \\ \hline
Avg.     & 90.10          & 86.21          & 89.93          & 88.67          & 84.16          & 67.04          & 
\end{tabular}
\caption{Patch-level accuracy of a six-class recognition model when trained on one alteration method and evaluated on all. In bold, performance on the alteration used for training. Last column: model average across all transformations. Bottom row: average performance of all models for each transformation.}
\label{tab:alteration_confusion}
\end{table}

Table~\ref{tab:alteration_confusion} shows blur is the transformation that most impacts detector performance. \textit{GaussianBlur}, which causes drops in accuracy of over 7 points, is also the hardest transformation in training, showing the lowest diagonal score. However, both blur-trained models achieve the highest cross-transformation accuracies, demonstrating effective generalization and making blur a valuable addition to the training process.



\section{Deployment Experiments} \label{sec:eval}

To study the impact of deployment factors on generalization, we use \justsusy{}, a multi-class model trained with the setup described in \S\ref{sec:train_experiments}. Training data from \S\ref{subsec:data} is augmented with all transformations from \S\ref{subsec:alterations}, each applied with a 20\% chance. Using this model generalization to images from new and external data sources is explored in \S\ref{subsec:source_eval}. The aggregation of patch-level SID predictions into image-level decisions is considered in~\S\ref{subsec:image_eval}. Finally, \S\ref{subsec:scale_generalization} studies the impact of input resolution changes on model generalization.



\subsection{Generalization to Source}\label{subsec:source_eval}

The \textit{\justsusy{} (Patch)} column of Table~\ref{tab:source} shows the results of evaluating \thesusy{} under disjoint sets of data (see \S\ref{subsec:data_eval}). For authentic datasets, three new sources of images are added, with varied results. The model generalizes well in \flickr{}, moderately in \glvtwo{} and poorly for \itw{} images.

\begin{table}[t]
\begin{tabular}{llcc}
\textbf{Data}   & \multicolumn{1}{l|}{\textbf{SIG}}   & \textbf{\justsusy} & \textbf{\justsusy} \\
\textbf{Source} & \multicolumn{1}{l|}{\textbf{Model}} & \textbf{(Patch)}   & \textbf{(Image)}   \\ \hline
\multicolumn{4}{l}{\textbf{Authentic Data Sources}}                                             \\ \hline
\flickr         & \multicolumn{1}{l|}{None}           & 91.81              & \textbf{94.48}     \\
\glvtwo         & \multicolumn{1}{l|}{None}           & 68.37              & \textbf{71.80}     \\
\itw            & \multicolumn{1}{l|}{None}           & \textbf{30.91}     & 27.27              \\ \hline
\multicolumn{4}{l}{\textbf{Synthetic Data Sources (Models in train)}}                           \\ \hline
\syn            & \multicolumn{1}{l|}{SD 1.3}         & 88.56              & \textbf{91.80}     \\
\syn            & \multicolumn{1}{l|}{SD 1.4}         & 88.50              & \textbf{91.80}     \\
\syn            & \multicolumn{1}{l|}{MJ V5}          & 74.22              & \textbf{78.40}     \\
\syn            & \multicolumn{1}{l|}{SD XL}          & 79.22              & \textbf{83.80}     \\
\syn            & \multicolumn{1}{l|}{DALLE-3}        & 87.02              & \textbf{92.50}     \\ \hline
\multicolumn{4}{l}{\textbf{Synthetic Data Sources (Models not in train)}}                       \\ \hline
\syn            & \multicolumn{1}{l|}{Glide}          & 52.78              & \textbf{53.20}     \\
\syn            & \multicolumn{1}{l|}{SD 2}           & 68.32              & \textbf{70.40}     \\
\syn            & \multicolumn{1}{l|}{DALLE-2}        & \textbf{24.50}     & 19.70              \\
\syn            & \multicolumn{1}{l|}{Firefly}        & 53.04              & \textbf{53.50}     \\
\texttt{Authors} & \multicolumn{1}{l|}{SD 3}           & 91.51              & \textbf{95.23}     \\
\texttt{Authors} & \multicolumn{1}{l|}{FLUX.1}         & 94.37              & \textbf{97.05}     \\
\itw            & \multicolumn{1}{l|}{Unknown}        & 90.51              & \textbf{91.92}    
\end{tabular}
\caption{Top: unseen authentic image datasets. Middle: unseen synthetic datasets produced by models seen during training. Bottom: synthetic datasets from unseen models. Recall at patch level and five-patch majority voting at image level. Best in bold.}
\label{tab:source}
\end{table}

The middle section of Table~\ref{tab:source} shows generalization on datasets from generators seen during the training stage (see \S\ref{subsec:data}). These datasets are from the same SIG models but generated by different users. Variations in SIG configurations, prompts and post-processing, may introduce significant biases. Nonetheless, generalization of \justsusy{} is good, reaching a recall between 74\% and 88\% in all cases.


The third set of experiments, reported at the bottom of Table~\ref{tab:source}, considers datasets generated by models unseen during training. Performance in this set has a large variance, with models reaching recalls as high as 94\% and as low as 24\%. The impact of model family on generalization is inconsistent: \thesusy{} excels on~\sdthree{}, performs adequately on~\sdtwo{} and struggles with~\dalletwo{}, despite being trained on versions of~\texttt{Stable Diffusion} and~\texttt{DALLE}. 

\subsection{Image Decision Boundary}\label{subsec:image_eval}


While SID models operate on small image patches, real-world applications typically require whole-image predictions. To address this gap, we analyze the top five patches selected based on texture complexity, as described in \S\ref{subsec:arch}.

We evaluated two aggregation strategies: majority voting of patch predictions and averaging patch logits before classification. Although both approaches demonstrate improvements over single-patch, majority voting consistently outperforms across datasets, with its results shown in the last column of Table~\ref{tab:source}. With this method, high-performing datasets showed further improvements, while poorly performing ones saw minimal gains, and those scoring below random chance experienced slight degradation. These findings highlight both the advantages and limitations of decision boundary tuning.

\begin{figure}[tb]
  \centering
  \includegraphics[width=0.98\linewidth]{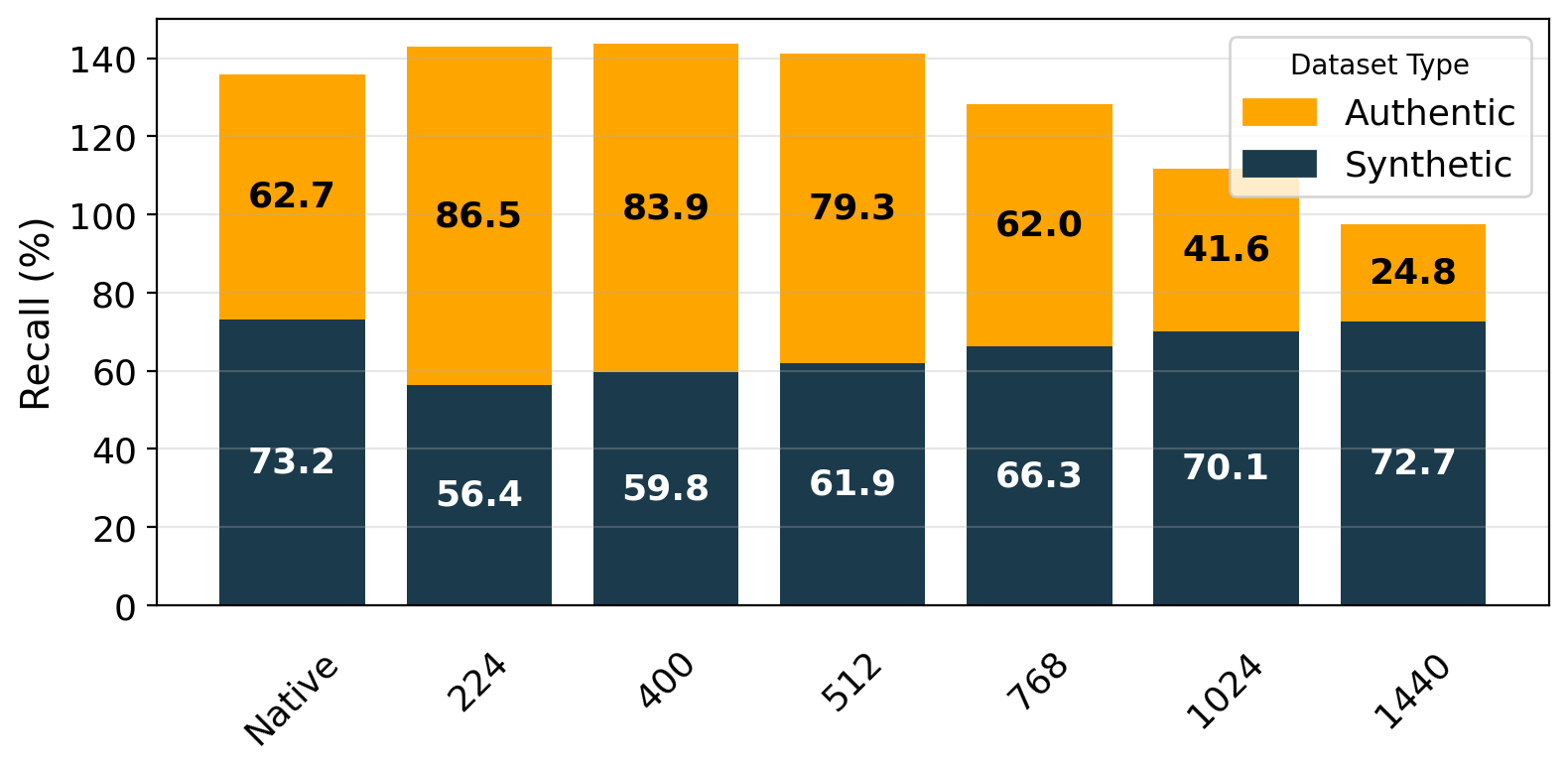}
  \caption{Recall of \susy{} on authentic and synthetic evaluation datasets, under different scaling factors.}
  \label{fig:susy_res}
\end{figure}

\subsection{Scale Generalization}\label{subsec:scale_generalization}



Image resizing is a widespread image alteration, and almost impossible to prevent. It can alter frequency artifacts and defects that SID models rely on, decreasing their performance. To assess the extent of this factor, we evaluate \thesusy{} using images scaled at six different sizes (224 to 1440). First, if the image is not already square, equal padding is added on both sides of the shorter dimension to center it. Then the squared image is resized to the specified dimensions using bilinear interpolation. Using the evaluation datasets described in \S\ref{subsec:data_eval}, which originally follow a diverse distribution of sizes (see \appx~\ref{appendix-resolution}), the recall for each dataset is computed individually (see \S\ref{subsec:arch}). Then, for each of the six image scales, the results for authentic and synthetic classes are averaged separately, allowing the monitoring of both accuracy and balance in detection. This experiment is reproduced in the benchmarking analysis of \S\ref{sec:generalization_sota}, for comparison with other SID models.


As shown in Figure~\ref{fig:susy_res}, \thesusy{} is not severely affected by resolution changes, only at higher rates. It achieves better combined results (around 140) at lower resolutions, with rescaling at 224, 400 and 512 being equally competitive. As resolution increases there is an increasing bias towards synthetic predictions. To ensure consistent performance in real-world applications, where images may have undergone prior resizing, we recommended including standardized rescaling in preprocessing pipelines.

\begin{table}[b]
\setlength\tabcolsep{4.3pt}
\centering
\begin{tabular}{c|cc}
\itw{}               & Authentic      & Synthetic      \\ \hline
Avg. Human Evaluator & 72.82          & 69.14          \\
\justsusy{} (Patch)  & 72.56          & 64.24          \\
\justsusy{} (Image)  & \textbf{75.21} & \textbf{69.70} \\
\end{tabular}
\caption{\itw{} recall by \thesusy{} and human evaluators, best in bold. For \justsusy{}, average performance at patch level, five-patch majority voting at image level.}
\label{tab:susy_vs_human}
\end{table}

\begin{figure*}[t]
  \centering
  \includegraphics[width=0.97\linewidth]{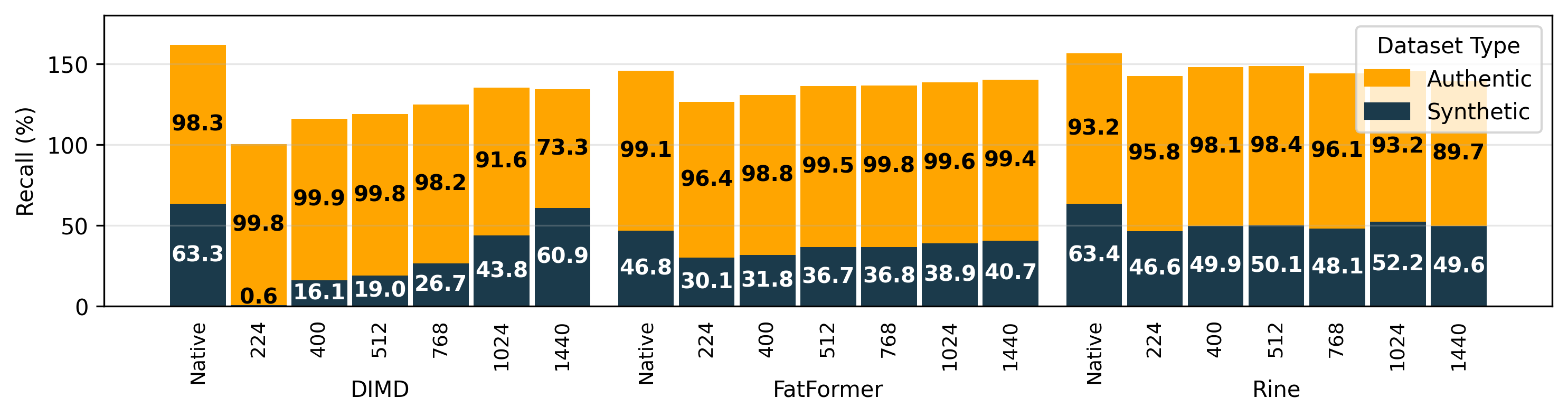}
  \includegraphics[width=0.97\linewidth]{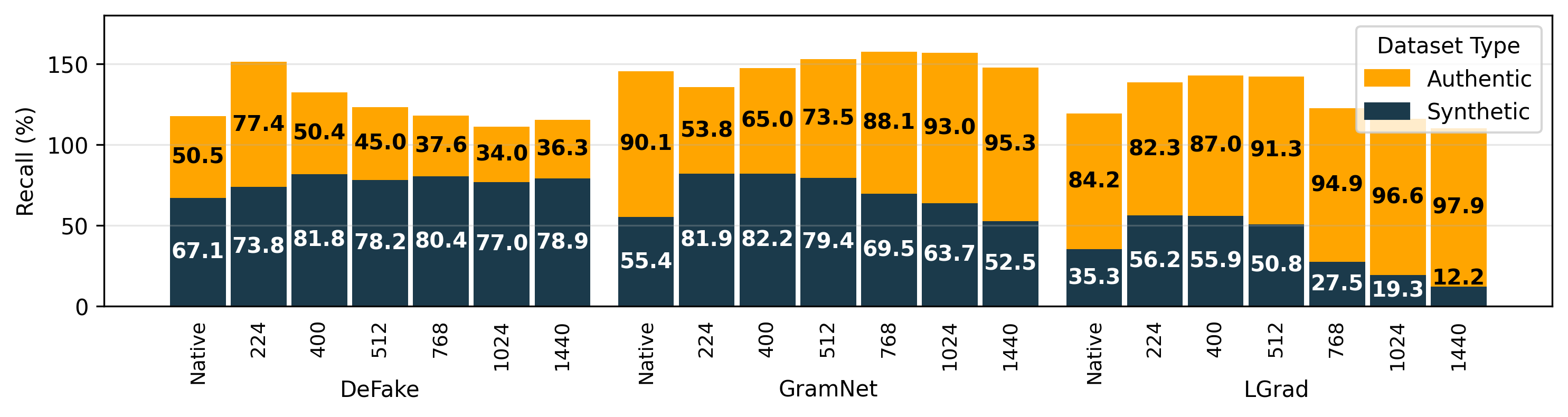}
  \caption{Recall of SID on authentic and synthetic evaluation datasets, under different scaling factors.}
  \label{fig:bench_res}
\end{figure*}

\subsection{Human Evaluators}

To measure the performance of SID models against human visual assessment, we use the \itw{} dataset, containing both authentic and synthetic images (see \S\ref{subsec:data_eval} for details). We ask 10 volunteers aged 22-30 who have social media accounts and are likely to be exposed to digital media and AI-generated content, to discriminate between both \itw{} versions (authentic and synthetic). To ensure unbiased results, images were presented in random order and the evaluators were not informed about the distribution of the data. All participants viewed the images on the same IPS LCD display (1920x1200 resolution, 400 nits brightness) in a controlled lighting environment. Participants took an average of 15 minutes to label the 210 images, with no time constraints imposed. Results are reported in Table~\ref{tab:susy_vs_human}. \susy{} outperforms the average human evaluator at image level, using the aggregation mechanism of \S\ref{subsec:image_eval} and the best resolution studied in \S\ref{subsec:scale_generalization}.

\section{Benchmarking Experiments}\label{sec:generalization_sota}

To complete this study on the generalization capacity of SID, we test the performance of ten different models (most made available through SIDBench~\cite{schinas2024sidbench}). Table~\ref{tab:sota} showcases the performance of the best six models (over 140 combined recall): LGrad~\cite{tan2023learning}, GramNet~\cite{liu2020global} and DIMD~\cite{koutlis2024leveraging}, which use CNNs as feature extractors, each with a unique emphasis on different image characteristics, together with Rine~\cite{koutlis2024leveraging}, DeFake~\cite{corvi2023detection} and FatFormer~\cite{liu2023fatformer}, based on transformer models. Further details on their architecture are given in \appx~\ref{appendix-metrics}, which also includes results for the other tested detectors: CNNDetect~\cite{wang2020cnn}, Dire~\cite{wang2023dire}, FreqDetect~\cite{frank2020leveraging}, and UnivFD~\cite{ojha2023towards}.



\subsection{Rescaling}\label{subsec:bench_scaling}

Given the crucial role of rescaling, the performance of current detectors is first assessed using the experimentation introduced in \S\ref{subsec:scale_generalization}, providing insights on the generalization of SID under scale changes and pointing towards the ideal setup for each detector. The models displayed in the top row of Figure~\ref{fig:bench_res} are highly sensitive to \textit{any} scale modifications, with their performance consistently deteriorating after rescaling (\ie optimal performance without resizing). This sensitivity to scale changes creates a security vulnerability that malicious actors could exploit, compromising the detectors' reliability. Moreover, these models demonstrate a notable bias toward the authentic class, achieving suboptimal recall scores for synthetic images (63\% for two models, while the third performs below random chance).

In contrast, the detectors shown in the bottom row of Figure~\ref{fig:bench_res} demonstrate resilience to \textit{some} scale variations (\ie optimal performance includes resizing). DeFake and LGrad perform optimally at lower resolutions (224, 400 or 512), similarly to \thesusy{} (see Figure~\ref{fig:susy_res}), while GramNet excels at higher resolutions (768 or 1024). Their optimal rescaled input resolution enhances their resilience and enables the optimization of deployment pipelines. However, these models differ significantly in their prediction balance: DeFake shows stronger performance in synthetic class detection, LGrad excels in authentic class identification, while GramNet and \susy{} achieve a more balanced performance across both categories.

\begin{table*}[t]
\setlength\tabcolsep{4.5pt}
\renewcommand{\arraystretch}{1.2}
\centering
\begin{tabular}{@{}llc!{\vrule width 1pt}ccccccc!{\vrule width 1pt}c@{}}
 & \textbf{SIG Model} & \textbf{Year} & {\small \textbf{DIMD}} & {\small \textbf{FatFormer}} & {\small \textbf{Rine}} & {\small \textbf{DeFake}} & {\small \textbf{LGrad}} & {\small \textbf{\justsusy{}}} & {\small \textbf{GramNet}} & \textbf{Avg.}\\ 
\midrule
\textbf{Best Resolution} & & & Native & Native & Native & 224 & 400 & 400 & 768 &  \\
\midrule
\multicolumn{10}{c}{\textbf{Authentic Data Sources}} \\  \midrule
\flickr{}              & -          & 2014 & 99.92 & \textbf{100.0} & 99.54 & 93.62 & 99.82 & 94.76 & 99.94 & \textbf{98.23} \\
\coco{}                  & -          & 2017 & \textbf{100.0} & 99.60 & \textbf{100.0} & 91.33 & 77.63 & - & 87.12 & 92.61 \\
\glvtwo{}          & -          & 2020 & 96.54 & 99.92 & 77.42 & 78.32 & 98.66 & 82.62 & \textbf{100.0} & 90.50 \\
\itw{}           & -          & 2024 & 96.69 & \textbf{96.77} & 95.87 & \underline{46.28} & 71.90 & 74.38 & 65.29 & 78.17 \\ \midrule \rowcolor{lightorange}
\textbf{Average Authentic}       &            &      & 98.29 & \textbf{99.07} & 93.21 & 77.39 & 87.00 & 83.92 & 88.09 & \\
\midrule
\multicolumn{10}{c}{\textbf{Synthetic Data Sources}} \\  \midrule
\syn{}           & Glide      & 2021 & \underline{6.10} & 68.10 & 83.60 & \textbf{86.50} & 53.50 & 53.30 & 68.80 & 59.99 \\
\mjtti{}                & MJ V1/V2    & 2022 & \underline{2.87} & 55.29 & \underline{14.57} & \textbf{75.83} & \underline{42.60} & - & 63.58 & \underline{42.46} \\
\syn{}            & SD 1.3     & 2022 & \textbf{100.0} & 88.20 & 99.90 & 86.20 & 81.10 & 87.00 & 90.00 & 90.34 \\
\syn{}            & SD 1.4     & 2022 & \textbf{100.0} & 88.00 & 99.60 & 87.20 & 81.30 & 87.10 & 90.70 & \textbf{90.56} \\
\ddb{}           & SD 1.X     & 2022 & \textbf{99.92} & 86.33 & 96.03 & 76.01 & 52.11 & - & 93.52 & 83.99 \\
\syn{}            & SD 2       & 2022 & \textbf{97.10} & \underline{47.20} & 85.80 & \underline{39.80} & 53.80 & \underline{42.30} & 75.50 & 63.07 \\
\syn{}                & DALLE-2          & 2022 & \underline{0.40} & \underline{45.80} & 70.80 & \underline{47.30} & 76.00 & \underline{20.70} & \textbf{93.10} & 50.59 \\
\syn{}            & MJ V5      & 2023 & \textbf{98.10} & \underline{30.60} & 87.00 & \underline{49.90} & 83.60 & \underline{36.50} & 88.50 & 67.74 \\
\mji{}             & MJ V5/V6    & 2023 & \textbf{90.11} & \underline{5.16} & \underline{31.28} & 75.85 & \underline{8.27} & - & \underline{15.56} & \underline{37.71} \\
\syn{}            & SDXL      & 2023 & 94.40 & 79.80 & \textbf{97.60} & 53.60 & 86.20 & \underline{45.90} & 97.20 
 & 79.24 \\
\realsdxl{}                  & SDXL       & 2023 & \textbf{97.65} & \underline{28.64} & 82.17 & 91.82 & 69.77 & - & 75.61 & 74.28 \\
\syn{}            & Firefly    & 2023 & \underline{18.10} & 81.40 & \underline{43.30} & 54.00 & 66.40 & \underline{24.50} & \textbf{83.00} & 52.96 \\
\syn{}            & DALLE-3    & 2023 & \underline{0.00} & \underline{0.00} & \underline{2.00} & \textbf{93.60} & \underline{35.00} & 84.30 & \underline{30.40} & \underline{35.04} \\
\dalle         & DALLE-3    & 2023 & 61.82 & \underline{3.92} & \underline{28.79} & \textbf{81.21} & \underline{3.03} & - & \underline{1.52} & \underline{30.05} \\
\texttt{Authors'} \sdthree{} & SD 3       & 2024 & \textbf{99.24} & 59.02 & 85.05 & 89.42 & 70.74 & 78.44 & 89.03 & 81.56 \\
\texttt{Authors'} \flux{} & Flux.1-dev & 2024 & 62.89 & \underline{23.08} & 54.72 & 81.43 & 63.92 & 85.40 & \textbf{92.99} & 66.35 \\
\itw{} & Unknown    &      & \underline{47.47} & \underline{5.00} & \underline{16.16} & \textbf{84.85} & \underline{22.22} & 71.72 & \underline{32.32} & \underline{39.96} \\ \midrule \rowcolor{lightblue}
\textbf{Average Synthetic}     &            &      & 63.30 & \underline{46.80} & 63.43 & \textbf{73.80} & 55.86 & 59.76 & 69.49 & 
\end{tabular}
\caption{Center-patch recall of detector models evaluated with their best input resize resolution. \textit{Native} denotes no resolution alteration. Top: Performance on authentic datasets. Bottom: Performance on synthetic datasets. Best recall in bold. Recalls below 50\% (worse than random) underlined. Entries denoted by (-) in~\justsusy{} indicate datasets excluded from evaluation as they were used for training.}
\label{tab:sota}
\end{table*}

\subsection{Optimal Model Generalization}

This final experiment evaluates the generalization capacity of existing detectors across benchmarking datasets after identifying their optimal input scales. Results in table~\ref{tab:sota} reveal a critical SID limitation: all detectors achieve less than 50\% recall on at least four datasets. This demonstrates no universal detector exists, as all methods eventually perform worse than random chance. Performance metrics averaged across dataset types (authentic vs synthetic) consistently favor the authentic class. While this bias partially stems from the selection of optimal resolution (\ie DeFake, GramNet and \susy{} all have input resolutions where synthetic class detection is higher, see Figure~\ref{fig:bench_res}), it also reflects the more diverse and challenging distribution within the synthetic class. Additionally, a clear trade-off emerges across dataset types: DeFake is simultaneously the best synthetic detector and worst authentic detector, while FatFormer shows the opposite pattern.

While DeFake is the best synthetic detector on average and DIMD excels in 8 out of 17 synthetic datasets (including 6 out of 7 StableDiffusion variants), both suffer from high sensitivity to rescaling (see \S\ref{subsec:bench_scaling}). This limitation makes them unsuitable for deployment scenarios with uncontrolled inputs. DIMD's consistent performance across \texttt{Stable Diffusion} models stands as an exception, as detectors generally show little consistency when handling different models within the same family. Performance varies even when two datasets are generated using the same model: the average recall difference between \dallethree{} versions across SID models is 27.63\%, while \sdxl{} variations average 24.35\%. While detectors may generalize to source changes under specific conditions (see \S\ref{subsec:source_eval}), this is not universal. Despite the \syn{} benchmarking datasets being generated using identical prompts and likely containing similar visual elements, detection performance varies substantially, as shown in Table~\ref{tab:sota}.


Private models, like \texttt{DALLE}, \texttt{Midjourney} and \firefly{}, present generalization challenges. Detectors achieve only 45.22\% average recall on closed SIG models, with the \textit{best} closed dataset reaching 67.74\%. In contrast, the same detectors achieve 76.60\% average recall on open SIG models, with even the \textit{worst} open dataset achieving 59.99\% recall. These findings underscore the crucial role of open science in advancing the field.



The \itw{} dataset, serving as a proxy for real-world performance, reveals additional limitations. No tested detector achieves above 50\% recall for \textit{both} authentic and synthetic versions across all input resolutions. Only \thesusy{} demonstrates robust performance, achieving over 70\% recall in both subsets, but specifically when operating at its optimal input resolution.

\section{Conclusions}\label{sec:concl}


In a race equilibrium paradox, better generative models appear regularly, making the task harder for humans, while detectors trained on these newer generators are more reliable (see \S\ref{sec:single-class}), keeping the race close.

The demand for detectors grows as society seeks to preserve social trust and digital rights while combating disinformation. Yet these detectors must improve their generalization capabilities to be truly effective. In that regard, the main lesson from this work is: \textit{never} assume generalization in SID. Results in Table~\ref{tab:sota} indicate even within datasets produced by the same generative model, detection performance may largely vary, as a result of software and hardware setups and user bias. Similarly, generalization should not be assumed on synthetic images produced by older, less realistic generators either, even if these synthetic samples seem more obvious to the human eye. As shown in Table~\ref{tab:single_class_recall_patch}, samples from these models are hard to generalize (but \textit{not} to train for) due to their stronger biases and distinct artifacts. In fact, even simple post-processing methods, like blur, can significantly reduce detector performance (see Table~\ref{tab:alteration_confusion}).

On top of this, image scale can dramatically affect the performance of most detectors, as well as the balance of their performance (\ie \textit{authentic} vs \textit{synthetic}). Some detectors are highly sensitive to rescaling operations (see Figure~\ref{fig:bench_res}), exposing a vulnerability to malicious inputs. At the same time, other SID models work optimally when applied to data that has been scaled to a certain size (see Figure~\ref{fig:bench_res}). This can be used to tune data for its detection, boosting performance on deployment settings (see Table~\ref{tab:sota}). 

The final contribution of this work, beyond the released \susy{}, code and datasets, is a list of policies for the SID field as a whole, including an ethical risk assessment. First, our work emphasizes the importance of openness in the field of generative AI. Results from Table~\ref{tab:sota} indicate open generative models can be more easily detected (+20\% combined recall points on average). While we are far from a universal detector (all detectors perform below random in some of our benchmarks), models trained for specific targets may be as good as humans at identifying synthetic content (see Table~\ref{tab:susy_vs_human}).




\subsection{Ethical Risks}\label{subsec:ethical}


Image detection systems pose significant ethical concerns, primarily due to their inherent fallibility. These systems produce both false positives and negatives (see Table~\ref{tab:sota}), potentially misidentifying authentic images as synthetic and vice versa. Such errors could infringe on digital rights and enable censorship. Therefore, human expert oversight is crucial when these systems are used in contexts affecting individual rights, and their outputs should never serve as definitive evidence.


Additionally, model bias remains a critical challenge. Training datasets often contain inherent biases that can skew detection results (\eg rural landscapes could be tagged as synthetic more often than urban images). Thorough evaluation across all relevant demographic and contextual factors is essential before deployment.

Furthermore, the datasets used for training may include samples with personal data. \coco~contains images of real people, and synthetic datasets used could include realistic depictions of specific individuals. However, given the training objective and parameter size of~\thesusy, it is highly unlikely that any such information could be encoded within the weights released in this work.

A final risk of releasing a SID model is dual use, as it can be used as a training objective for generative models (\eg adversarial training). To mitigate that, we add a specific clause in the terms of use of the model prohibiting such practice. Notice \thesusy{} is not trained to be the best possible detector (not trained on all data), and should not be used \textit{as is} in practice. We recommend any SID model produced for final use to be kept private, as long as its public release holds no special academic or social value.

\subsection{Future Work}\label{subsec:future-work}



The results of this work point towards four research directions that could improve SID robustness and adaptability. The complementary strengths of different detector models indicate potential benefits from ensemble methods that combine them. Exploring training data scaling laws could reveal further insights into data requirements and generalization capabilities. Given the impact of input resolution, developing multi-resolution architectures could provide inherent resilience against scaling-based evasion attempts. Lastly, extending detection capabilities to video content is crucial to address the increasing quality of video generation models. These advancements are critical to ensure SID keeps pace in the ongoing race with SIG.


\section*{Acknowledgements}
This work has been partly funded by the AI4Media and AI4Europe projects from the European Union's Horizon 2020 programme (Grant Agreements Nº951911 and Nº101070000), and by a SGR-GRE grant from the Generalitat de Catalunya (code 2021 SGR 01187). The authors would like to acknowledge Mauro Achile, Eric Arean, Nura Mangado, Diego Rios and Daniel Pulido who contributed to motivating and contextualizing this work. Special thanks to the volunteers who participated in the human evaluation experiment.

{
    \small
    \bibliographystyle{ieeenat_fullname}
    \bibliography{main}
}

\clearpage
\setcounter{page}{1}
\maketitlesupplementary
\appendix

\section{Model Architecture}\label{appendix-arch}

\begin{figure}[b]
  \centering
  \includegraphics[width=\linewidth]{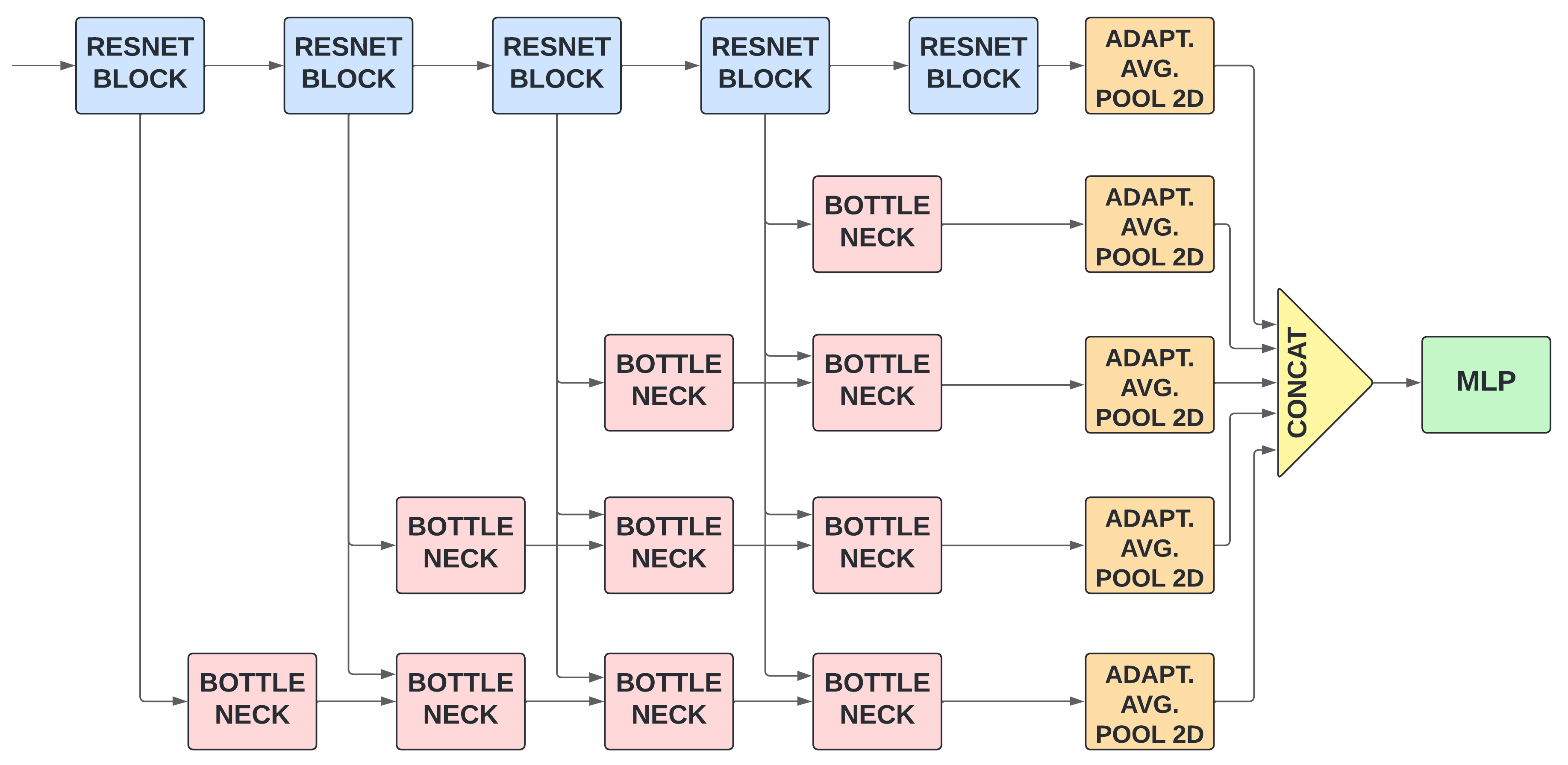}
  \caption{Detector architecture used, based on a ResNet-18 from~\cite{lopez2023sr}, including ResNet blocks (blue), bottlenecks (red), adaptative average pooling 2D (orange), concatenation (yellow) and an MLP (green).}
  \label{fig:arch}
\end{figure}

Figure~\ref{fig:arch} shows the detector architecture used for \susy{}, based on the design proposed in~\cite{lopez2023sr}. The architecture combines CNN-based feature extraction with MLP classification in a staircase design. The model employs a ResNet-18~\cite{he2016deep} backbone for feature extraction, totaling 12.7M parameters, with the CNN feature extractor accounting for 12.5M parameters and the MLP classifier using 197K parameters.

The CNN feature extractor implements five sequential stages following the ResNet-18 architecture. Each stage's output feeds into specialized bottleneck modules arranged in a staircase pattern. These bottleneck modules consist of three 2D convolutional layers that process and refine the extracted features. The staircase architecture creates a hierarchical feature processing system where each bottleneck level processes features from progressively later stages. The bottleneck modules combine inputs from their current stage and previous bottleneck module, except for the first module in each level. The staircase design enables the model to leverage features extracted at multiple depths, enhancing its ability to detect and classify synthetic content.

The classification component processes features through several steps. First, a 2D adaptive average pooling is applied to each bottleneck level output and stage 4. These pooled features are then concatenated to create a unified feature map. This feature map feeds into a three-layer MLP with dimensions of 512, 256 and 256 units, with dropout layers (at 0.5 probability) inserted between MLP layers to prevent overfitting.

\section{Experiment Setup}\label{appendix-setup}

Experiments were conducted on the MareNostrum 5 supercomputer, hosted at the Barcelona Supercomputing Center (BSC). We utilize an Intel Xeon Platinum 8460Y processor and one NVIDIA Hopper H100 64GB GPU. Seventy-five training runs were conducted with this setup, totalling sixteen hours of computing time, while continuously monitoring GPU power usage. Using the European Union's latest $\mathrm{CO_2}$ emission ratio~\cite{EEA_GHG_Emission_Intensity}, we estimated the carbon footprint of these experiments to be 0.63 kg of $\mathrm{CO_2}$.

\begin{figure}[t]
  \centering
  \includegraphics[width=\linewidth]{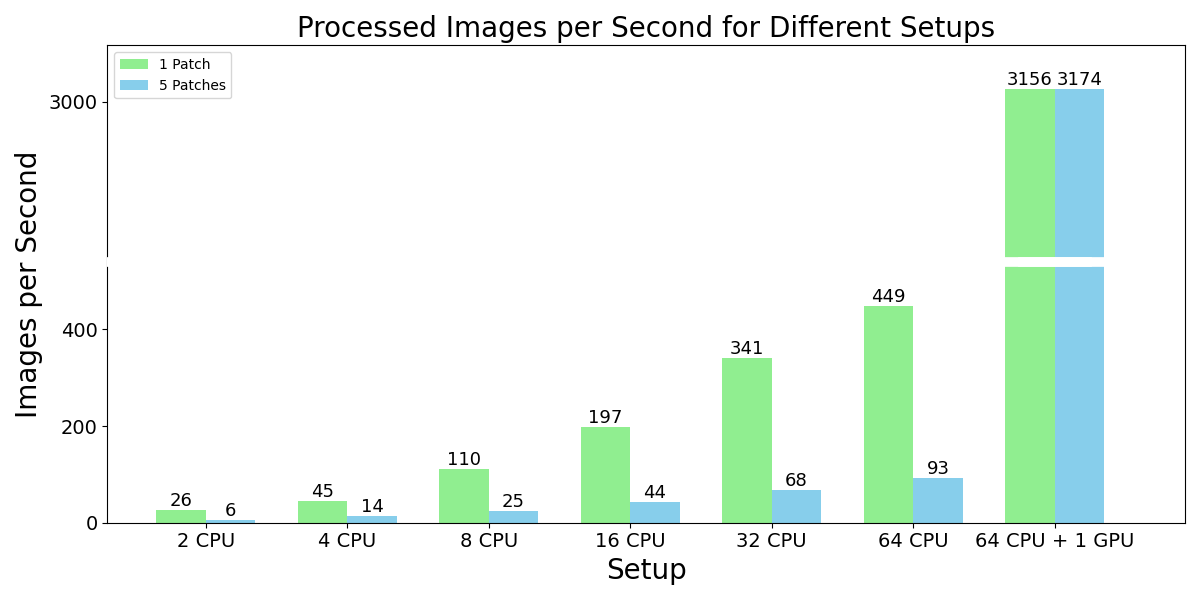}
  \caption{Scalability analysis showing images processed per second using single-crop (green) and 5-crop (blue) approaches across different hardware configurations. Note the axis break highlighting the GPU acceleration gain.}
  \label{fig:scalability}
\end{figure}

Figure~\ref{fig:scalability} presents a scalability analysis across different hardware configurations, ranging from 2 to 64 CPU cores, plus a hybrid setup combining 64 CPU cores with GPU acceleration. The evaluation compares processing speeds using single-crop and 5-crop approaches, measuring only the network's forward pass time. The results demonstrate near-linear scaling with CPU cores, with the single-crop approach consistently outperforming the 5-crop strategy in CPU-only configurations due to its lower computational requirements. However, this performance gap becomes negligible in the hybrid CPU-GPU setup, where both approaches achieve similar throughput of approximately 3,000 images per second, caused by the GPU's superior parallel processing capabilities for matrix operations. The improvement in processing speed with GPU acceleration, 7 to 34 times speedup compared with the fastest CPU-only setups, is highlighted by the broken y-axis in the plot.

\section{Training Datasets Details}\label{appendix-train-data}

\begin{table*}[tb]
\centering
\setlength\tabcolsep{5pt}
\begin{tabular}{cccccrrr}
\textbf{Train Dataset}      & \textbf{Model} & \textbf{Year} & \textbf{Image Format} & \textbf{Type} & \textbf{Train} & \textbf{Validation} & \textbf{Test} \\ \hline
\coco        & -           & 2017 & JPG  & Authentic & 2,967 & 1,234 & 1,234  \\
\dalle       & \dallethree & 2023 & JPG  & Synthetic & 987   & 330   & 330    \\
\ddb         & \sdone      & 2022 & PNG  & Synthetic & 2,967 & 1,234 & 1,234  \\
\sdxl        & \realsdxl   & 2023 & PNG  & Synthetic & 2,967 & 1,234 & 1,234  \\
\mjtti       & \mjearly    & 2022 & PNG  & Synthetic & 2,718 & 906   & 906    \\
\mji         & \mjlate     & 2023 & JPG  & Synthetic & 1,845 & 617   & 617    \\ \hline
\textbf{Evaluation Dataset} & & & & & & & \\ \hline
\flickr      & -           & 2014 & JPEG & Authentic & -     & -     & 31,655 \\
\glvtwo{}    & -           & 2020 & JPEG & Authentic & -     & -     & 5,000  \\
\itw         & -           & 2024 & Mix  & Authentic & -     & -     & 121    \\
\syn         & Many        & 2024 & PNG  & Synthetic & -     & -     & 9,000  \\
\sdthree     & \texttt{SD 3} & 2024 & PNG  & Synthetic & -     & -     & 8,192  \\
\flux        & \texttt{FLUX.1-dev} & 2024 & PNG  & Synthetic & -     & -     & 8,192  \\
\itw         & ?           & 2024 & PNG  & Synthetic & -     & -     & 99    
\end{tabular}
\caption{Datasets, including generative models included, release date, image format, authentic or synthetic, and number of samples within train, validation and test.}
\label{tab:dataset_splits}
\end{table*}

The \dalle{}~\cite{dalle-3-images} dataset contains 1,647 unique, deduplicated images generated by \dallethree{}~(2023), encompassing both photorealistic and digital art styles. Another dataset, the \ddb{} \cite{wangDiffusionDBLargescalePrompt2022}, was created using models of the 1.x Stable Diffusion series, which were released in 2022. In this dataset, we filter samples making sure that '\textit{photo}' appears in the prompt. Images from this dataset are of lower quality and visual detail than those of its successor \sdxl{}~\cite{sdxl}, which was released in 2023. The associated dataset, \sdxl{}~\cite{realisticSDXL} contains 5,435 images in the '\textit{realistic}' subset.

Beyond DALL-E and Stable Diffusion, the third main provider of synthetic images is Midjourney. Its early iterations, the V1 and V2 models, date from early 2022, and were used to populate the \mjtti{}~\cite{iulia_turc_gaurav_nemade_2022} dataset, which contains 4,530 images. Collage images and mosaics made of synthetic images were removed from this dataset. Later models, the V5 and V6 models from 2023 compose our last training dataset, \mji{}~\cite{midjourney-images}, with 1,226 images. This dataset also had to be deduplicated.

\section{Generated Datasets} \label{appendix-generated}

The input prompts used to produce the samples of \sdthree{} and \flux{} are extracted from \textit{Gustavosta/Stable-Diffusion-Prompts}\cite{Gustavosta2023}. For each image, height and width were randomly selected from a uniform distribution over the set $\{512, 768, 1024, 1344\}$ pixels. The images were generated using the official models~\cite{StabilityAI2024SD3, BlackForestLabs2024} accessed through HuggingFace. We employed a consistent generation process across all images, utilizing 28 inference steps for each generation. To enhance the quality and realism of the generated images, we added a set of negative prompts: \textit{poorly rendered face, poor facial details, poorly rendered hands, low resolution, blurry image, oversaturated, extra fingers, missing arms, missing legs, extra arms, extra legs, fused fingers, too many fingers, long neck, extra foot}. For \texttt{FLUX-1.dev}, inference is run using \textit{torch.bfloat16} precision, a guidance scale of 3.5 and a maximum sequence length of 512.

\begin{figure*}[tb]
  \centering
  \includegraphics[width=0.90\linewidth]{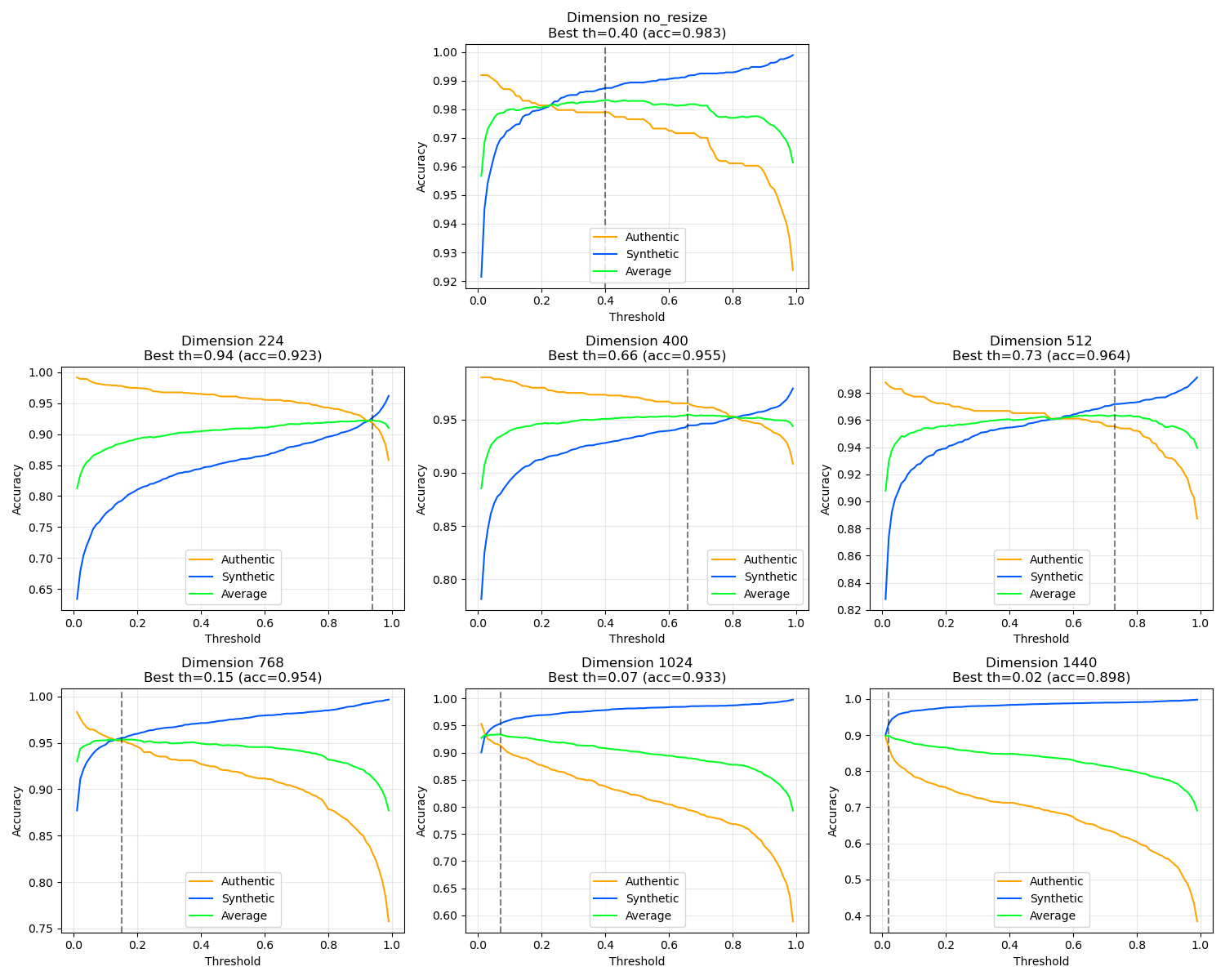}
  \caption{Classification accuracy curves for authentic (yellow), synthetic (blue) and green (average) test splits of \S\ref{subsec:data} datasets across seven resize dimensions.}
  \label{fig:th_test}
\end{figure*}

\section{Threshold Study} \label{appendix-threshold}

To transform the six-class model predictions into binary classifications, a threshold-based approach is tested as an alternative to the traditional \textit{argmax} method. This approach compares the probability score of the authentic class against a threshold value. Specifically, if the predicted probability for the authentic class is greater or equal than the threshold, the image is classified as authentic; conversely, if the probability falls below the threshold, the image is labeled as synthetic.

In the top part of Figure \ref{fig:th_test}, the performance of \thesusy{} with different thresholds is reported. The best performance is obtained with the threshold at 0.4, with the crossover point between authentic and synthetic accuracies being just above 0.2. Additionally, the threshold mechanism is tested with the different scale sizes of \S\ref{subsec:scale_generalization}, as reported in the other plots of figure \ref{fig:th_test}. With the scale changes, the crossover point moves progressively from above 0.9 at the smallest scale to approximately 0.01 at the largest scale. This behaviour mirrors the progression in class performance in Figure \ref{fig:susy_res}.

\begin{figure*}[tb]
  \centering
  \includegraphics[width=0.90\linewidth]{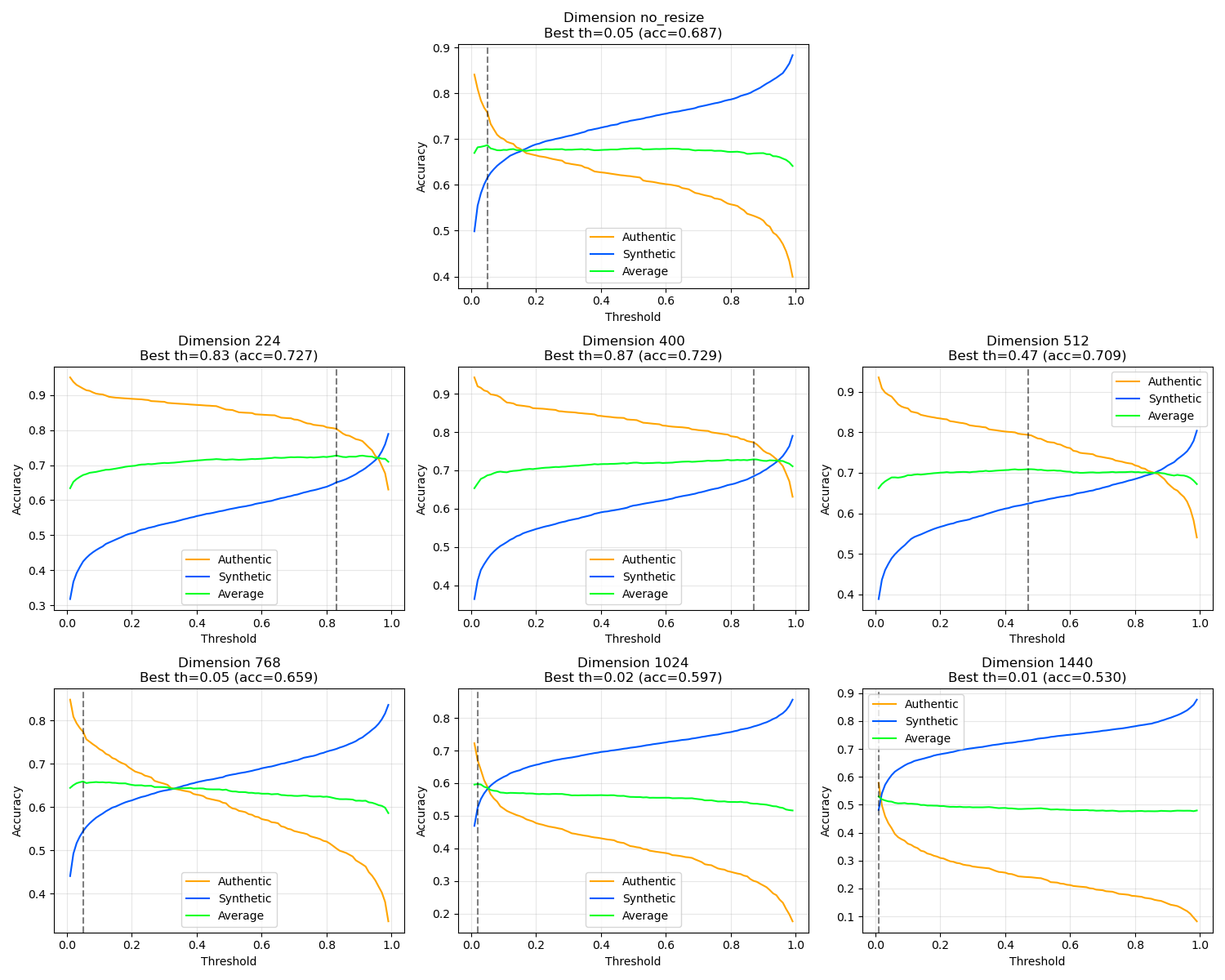}
  \caption{Classification accuracy curves for authentic (yellow) and synthetic (blue) data across seven resize dimensions, showing optimal threshold values and corresponding accuracy scores. The average performance (green) reveals that higher resize dimensions lead to lower optimal threshold values.}
  \label{fig:th_external}
\end{figure*}

We extended our threshold analysis to the external datasets described in \S\ref{subsec:data_eval}, as shown in Figure \ref{fig:th_external}. The results reveal a similar inverse relationship between image dimensions and optimal threshold values. However, these experiments uncovered more pronounced performance disparities between authentic and synthetic classifications compared with the previous experiment.

\section{Detector Models for Benchmarking} \label{appendix-metrics}


\textit{LGrad} trains a ResNet-50 classifier using image gradients from a pre-trained CNN, with images generated by ProGAN and authentic images from Celeba-HQ~\cite{karras2017progressive}. Similarly, \textit{GramNet} employs global image texture representations extracted at different levels from a ResNet-18, trained on StyleGAN-generated images and authentic Celeba-HQ images. \textit{Rine} leverages image representations extracted by intermediate blocks of CLIP, with an additional trainable module. We use the checkpoint trained with Latent Diffusion Model~\cite{corvi2023detection} and ProGAN images. \textit{DIMD} trains a ResNet-50 avoiding downsampling step, to preserve high-frequency fingerprints. We take the checkpoint trained on Latent Diffusion images. Training authentic images are taken from MSCOCO and LSUN. In \textit{DeFAKE}, text and image encoders from a Visual-Language Model model are finetuned to detect synthetic images, using Latent Diffusion data. \textit{Dire} uses the error between an input image and its reconstruction by a pre-trained diffusion model for identification. A ResNet50 is trained as a classifier on their DiffusionForensics dataset. Synthetic images generated with, ADM~\cite{dhariwal2021diffusion}, iDDPM~\cite{nichol2021improved} and PNDM~\cite{liu2022pseudo}, from LSUN-Bedroom~\cite{yu2015lsun} and Imagenet~\cite{deng2009imagenet}.
They hypothesize that diffusion-generated images can be approximately reconstructed by a diffusion model while real images cannot. \textit{FatFormer}~\cite{liu2023fatformer} adapts a pre-trained CLIP model by adding custom forgery-aware adapters to the image encoder to capture both low-level forgery artifacts and forgery traces in different frequency bands. It uses language-guided alignment, which leverages contrastive objectives between image features and text prompts.

\begin{table*}[ht]
\setlength\tabcolsep{4.5pt} 
\renewcommand{\arraystretch}{1.2} 
\centering
\begin{tabular}{@{}llcccccc@{}}
                      & \textbf{Model}      & \textbf{Year} & \textbf{CNNDetect}       & \textbf{FreqDetect} & \textbf{UnivFD}  & \textbf{Dire} \\ 
\midrule
\textbf{Best Resolution} & & & Native & 768 & 1024 & 224  \\ \midrule
\multicolumn{7}{c}{\textbf{Authentic Data Sources}} \\  \midrule
\flickr{}              & -          & 2014 & 98.28     & 98.80      & 93.26  & 65.38 \\
\coco{}                  & -          & 2017 & 98.14     & 97.89      & 94.98  & 66.45 \\

\glvtwo{}          & -          & 2020 & 98.54     & 99.62      & 93.72  & 73.08 \\
\itw{}           & -          & 2024 & 92.56     & 95.87      & 98.35  & 67.77 \\
\midrule \rowcolor{lightorange}
\textbf{Average Authentic}       &            &      & 96.88     & 98.05      & 95.08  & 68.17 \\
\midrule
\multicolumn{7}{c}{\textbf{Synthetic Data Sources}} \\  \midrule
\syn{}           & Glide      & 2021 & 17.10     & 76.70      & 60.70  & 26.10 \\
\mjtti{}                & MJ V1/2    & 2022 & 5.52      & 5.52       & 58.61  & 28.92 \\
\syn{}            & SD-1.3     & 2022 & 7.50      & 1.80       & 66.10  & 76.10 \\
\syn{}            & SD-1.4     & 2022 & 8.90      & 1.30       & 64.40  & 76.40 \\
\ddb{}           & SD 1.X     & 2022 & 9.72      & 6.24       & 56.48  & 28.20 \\
\syn{}            & SD-2       & 2022 & 23.50     & 0.80       & 43.80  & 54.90 \\
\syn{}                &  DALLE-2          & 2022 & 18.00     & 6.80       & 72.70  & 49.00 \\
\syn{}            & MJ-V5      & 2023 & 11.40     & 2.10       & 1.90   & 29.40 \\
\mji{}             & MJ V5/6    & 2023 & 0.65      & 1.78       & 6.16   & 31.12 \\
\syn{}            & SD-XL      & 2023 & 35.90     & 0.90       & 29.30  & 39.10 \\
\sdxl{}                  & SDXL       & 2023 & 18.40     & 6.16       & 4.21   & 27.88 \\
\syn{}            & Firefly    & 2023 & 20.80     & 6.20       & 33.80  & 38.20 \\
\syn{}            & DALLE-3    & 2023 & 0.10      & 0.70       & 0.20   & 69.90 \\
\dalle         & DALLE-3    & 2023 & 1.21      & 0.91       & 12.12  & 43.94 \\
Authors               & SD-3       & 2024 & 10.89     & 12.57      & 6.70   & 28.56 \\
Authors               & Flux.1-dev & 2024 & 5.73      & 2.89       & 1.83   & 32.37 \\
\itw{} & Unknown    &      & 7.07      & 3.03       & 2.02   & 50.51 \\ \midrule \rowcolor{lightblue}
\textbf{Average Synthetic}     &            &      & 11.91     & 8.02       & 30.65  & 42.98
\end{tabular}
\caption{Center-patch recall of other studied detector models across evaluation datasets. Each model is evaluated on the best resolution. Top: Performance for authentic images. Bottom: Performance on synthetic datasets. Recalls below 50\%, worse than random chance, underlined. Best recall for each dataset in bold.}
\label{tab:sota_discarded}
\end{table*}

The evaluation results in Table~\ref{tab:sota_discarded} reveal significant performance disparities across models when tested on authentic and synthetic datasets applying a center crop of size $224\times224$ to the input image. For \textit{CNNDetect}, while it achieves high recall rates on authentic datasets (96.88\% on average), its performance on synthetic datasets is poor, with an average recall of only 11.91\%. This suggests a strong reliance on features specific to GAN-based models in the training set, which are largely absent in modern diffusion-based architectures. Similarly, \textit{FreqDetect}~\cite{frank2020leveraging} demonstrates recall values close to 100 for authentic datasets but fails to adapt to the frequency artifacts produced by diffusion-based generators. \textit{UnivFD} shows more balanced performance across some diffusion-based datasets, suggesting better generalization. However, it still lags behind on more recent synthetic datasets. Dire stands out with notably low recall rates on both authentic (68.17\%) and synthetic (42.98\%) datasets. While it outperforms other models on a few synthetic datasets, its overall performance remains inconsistent.

\section{Benchmark Image Resolution Distribution} \label{appendix-resolution}

\begin{figure*}[h!]
    \centering
    \includegraphics[width=0.90\linewidth]       {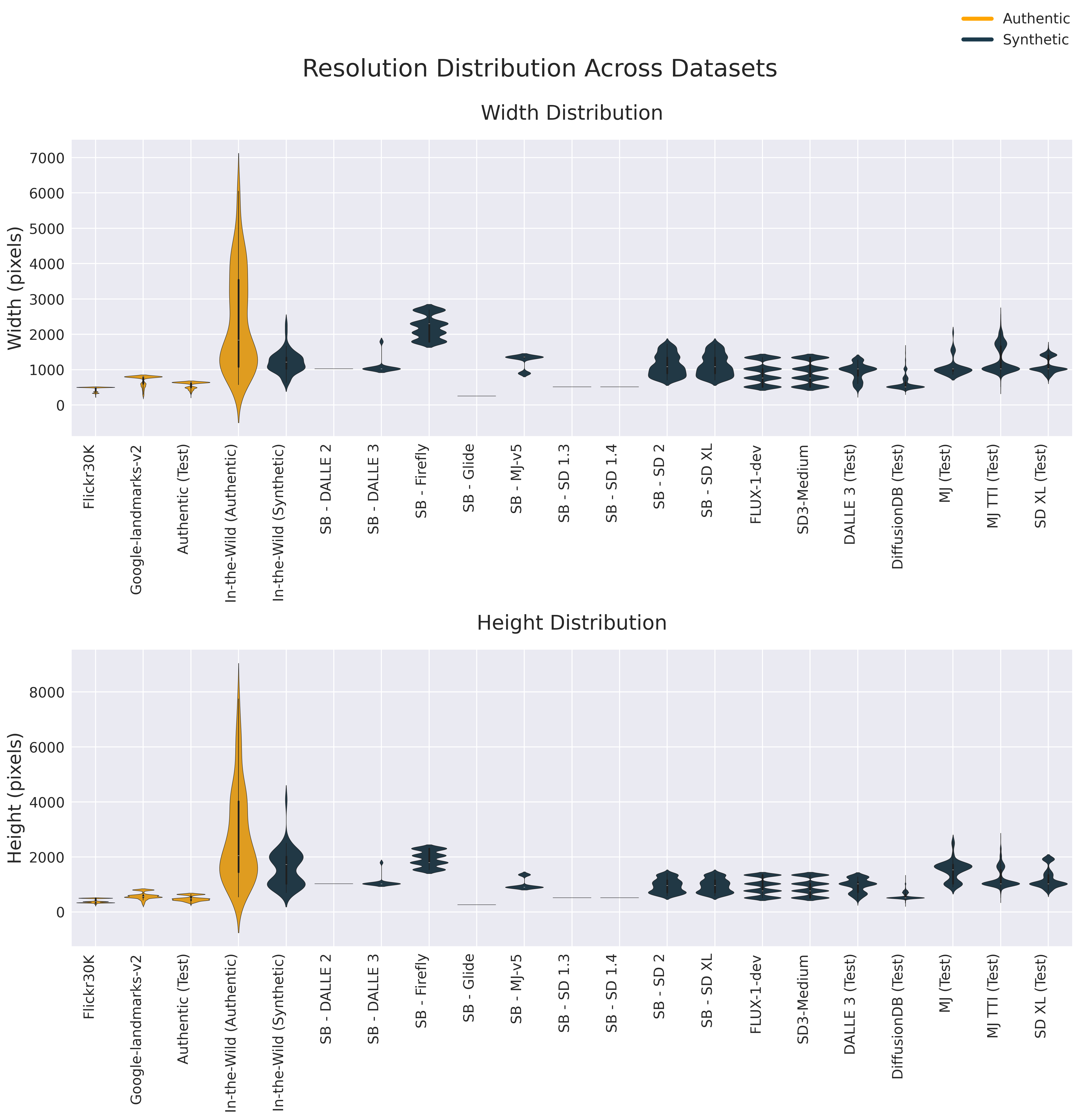}
    \caption{Resolution distribution of images across various datasets. The plots display the distribution of image widths (top) and heights (bottom). \texttt{SB} indicates datasets sourced from Synthbuster, \texttt{(test)} refers to datasets derived from our test splits.}
    \label{fig:resolution_distribution}
\end{figure*}

We calculate the resolution distribution for each of the evaluation datasets. Figure~\ref{fig:resolution_distribution} contains information regarding the size and aspect ratio of the datasets used. The top plot shows the width distribution of all samples, dataset-wise. The bottom plot shows the same information for height.

Figure \ref{fig:resolution_distribution} highlights distinct differences in resolution distribution between authentic and synthetic datasets. Authentic datasets, such as \itw{}, exhibit a broad range of resolutions, with widths and heights reaching up to 7000 and 8000 pixels, respectively, reflecting real-world variability. Synthetic datasets demonstrate narrower resolution distributions that are similar to each other. Among synthetic datasets, \texttt{Firefly} contains the images with the highest resolutions.

\section{Benchmarking Dataset Image Samples} \label{appendix-sample-images}

\begin{figure*}[h]
  \centering
  \includegraphics[width=0.95\linewidth, trim={185px 0 0 0}, clip]{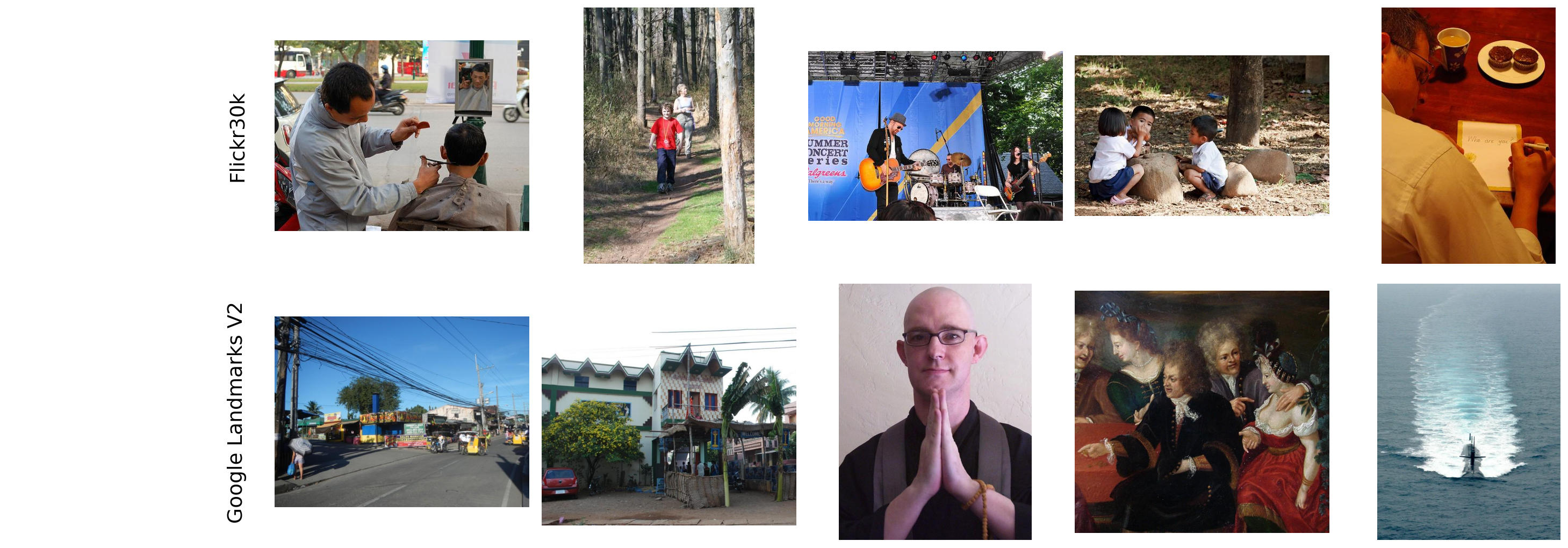}
  \caption{Sample authentic images from \flickr{} (top) and \glvtwo{} (bottom).}
  \label{fig:gl3}
\end{figure*}
\begin{figure*}[h]
  \centering
  \includegraphics[width=0.95\linewidth, trim={185px 0 0 0}, clip]{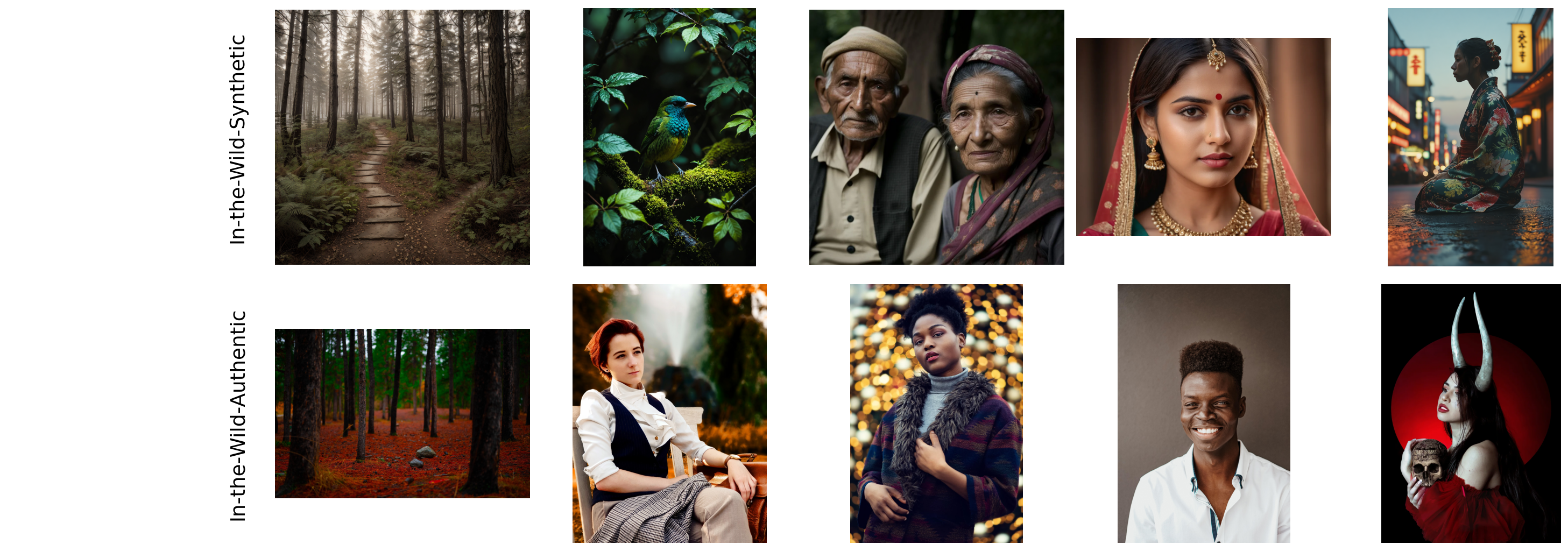}
  \caption{Sample images from our \itw{} dataset. Synthetic images (top) and authentic images (bottom)}
  \label{fig:itw_samples}
\end{figure*}
\begin{figure*}[hbt!]
  \centering
  \includegraphics[width=\linewidth, trim={185px 0 0 1250px}, clip]{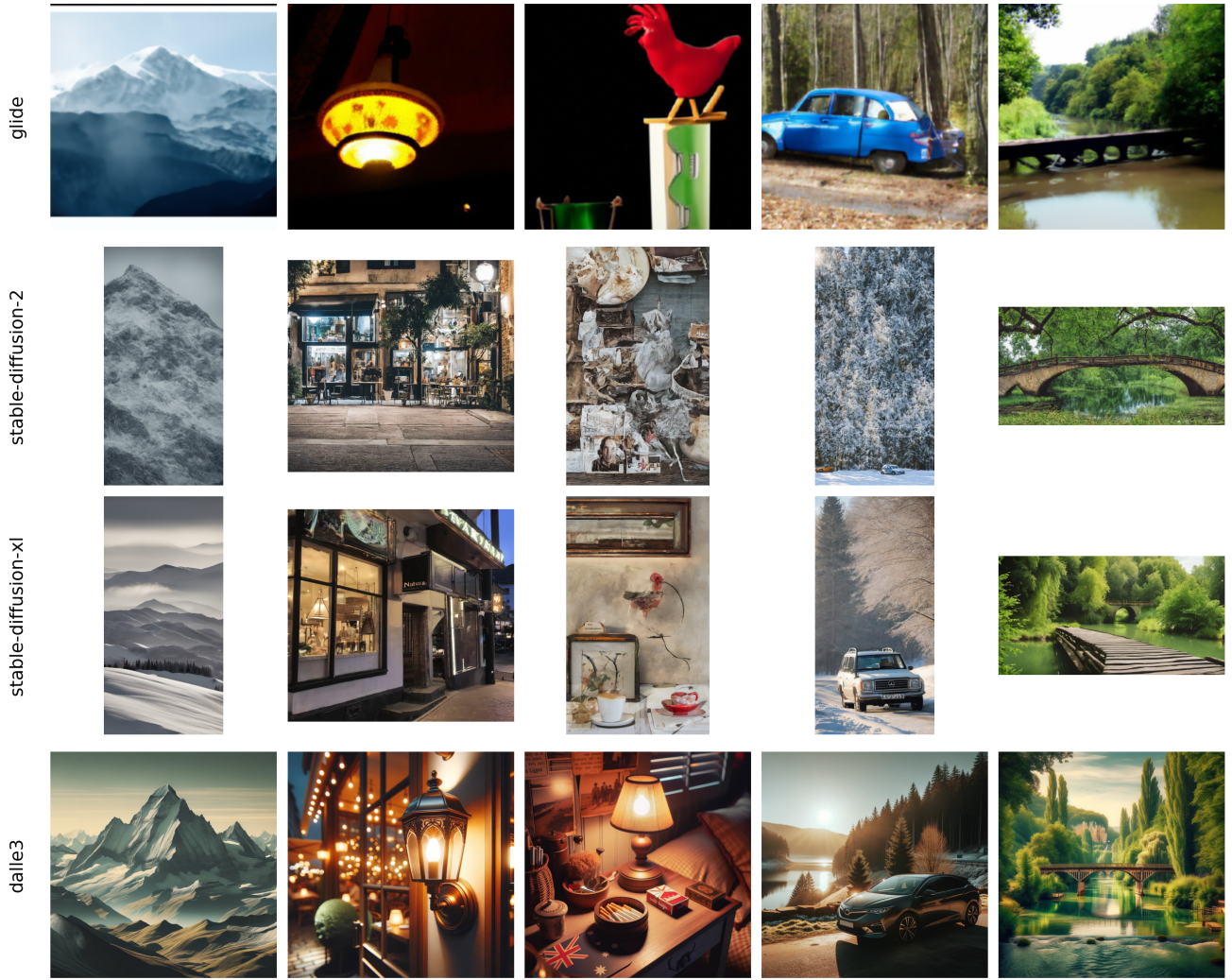}
  \caption{Sample synthetic images from \syn{}. Note that the same prompt is used for each column.}
  \label{fig:synth_samples}
\end{figure*}
\begin{figure*}[h]
  \centering
  \includegraphics[width=\linewidth, trim={185px 0 0 0}, clip]{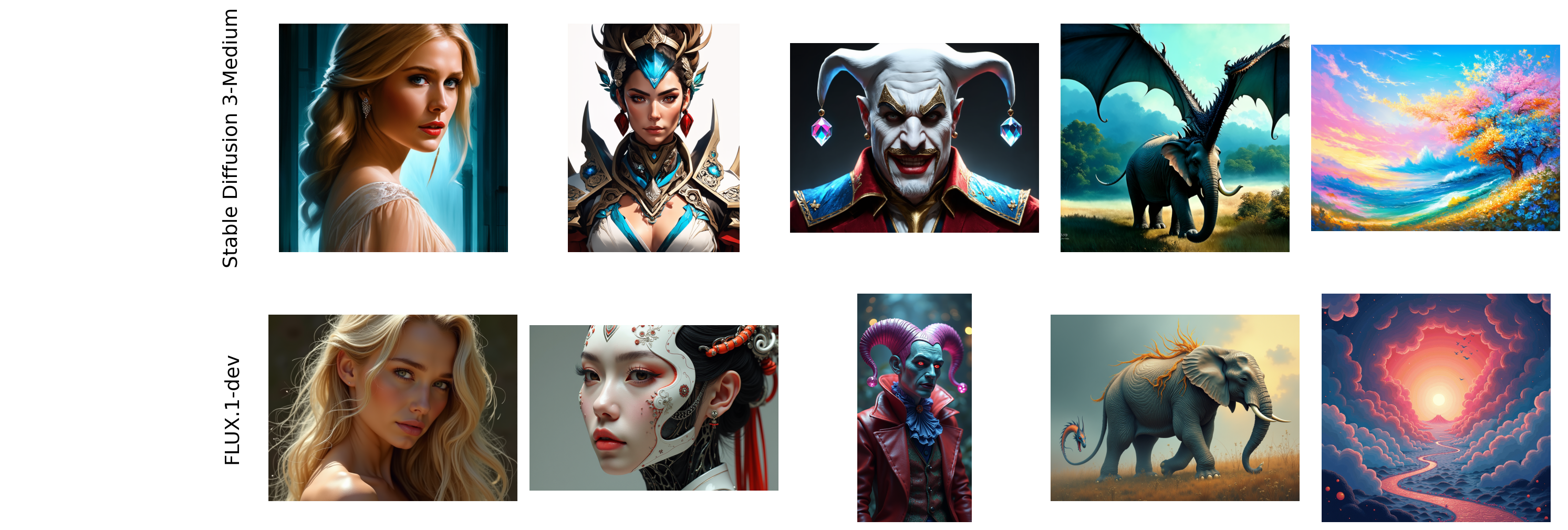}
  \caption{Sample synthetic images from our generated Stable Diffusion 3-Medium (top) and FLUX.1-dev (bottom) datasets.}\label{fig:sd3-flux}
  \label{fig:sd3-flux_samples}
\end{figure*}

This section includes several sample images from the evaluation datasets utilized:
\begin{itemize}
\item Figure~\ref{fig:gl3}: \glvtwo{} and \flickr{}.
\item Figure~\ref{fig:itw_samples}: \itw{}
\item Figure~\ref{fig:synth_samples}: \syn{} (\glide{}, \sdonethree{}, \sdonefour{}, \texttt{MJ 5}, \firefly, \dalletwo{}, \glide{}, \sdtwo{}, \sdxl{}, \dallethree{}).
\item Figure~\ref{fig:sd3-flux_samples}: \sdthree{} and \flux{}.
\end{itemize}


\begin{figure*}[h]
  \centering
  \includegraphics[width=\linewidth, trim={185px 1000px 0 0}, clip]{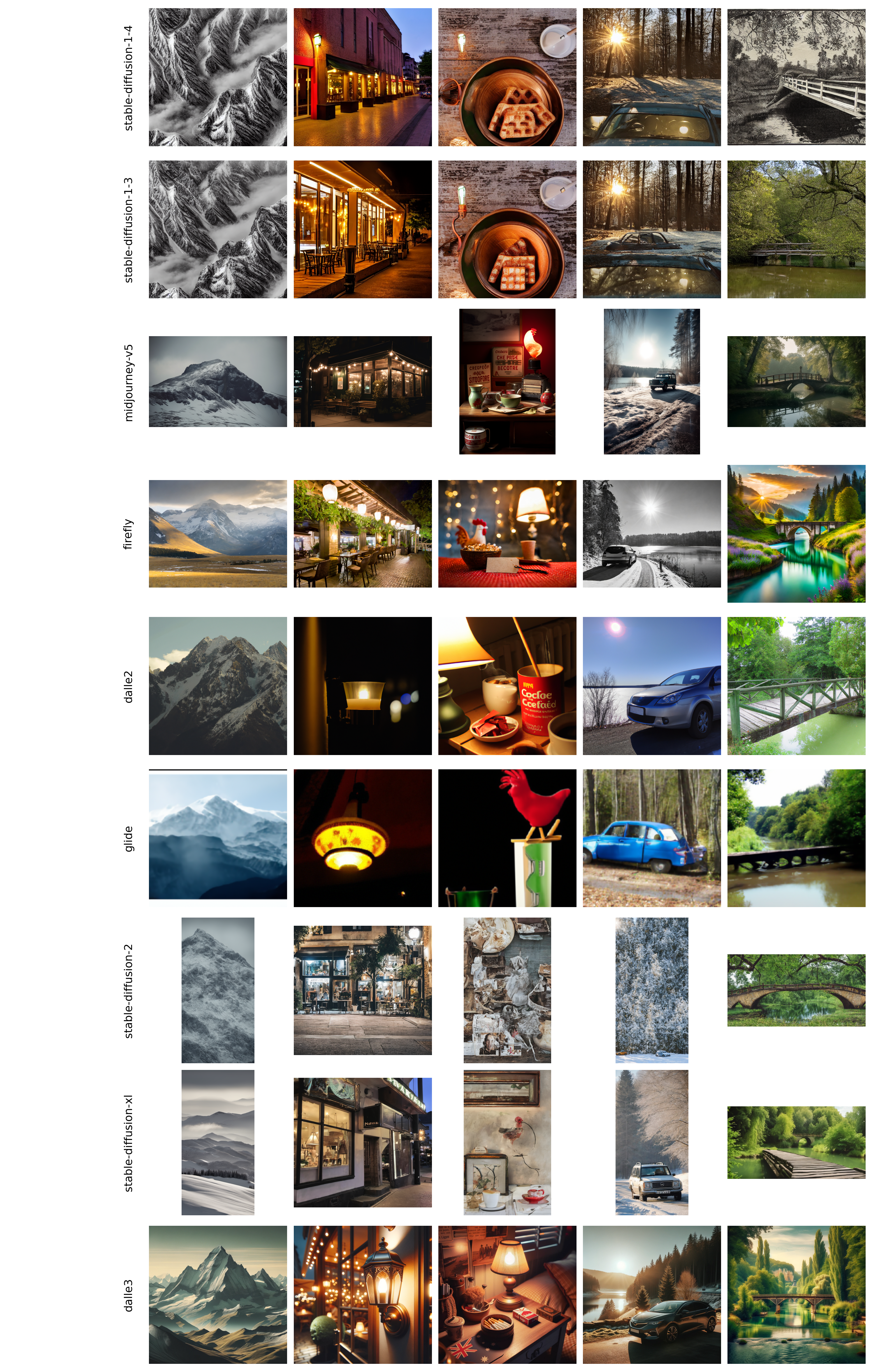}
\end{figure*}

\end{document}